\documentclass[times,twocolumn,final]{elsarticle}

\usepackage{medima}
\usepackage{framed,multirow}

\usepackage{latexsym}

\usepackage{amsmath}
\usepackage{multirow}
\usepackage{longtable}
\usepackage{amssymb}
\usepackage{lipsum}
\usepackage{subfigure}
\usepackage{graphicx}
\usepackage{pifont}
\usepackage{amsmath}
\usepackage{amsthm}
\usepackage{makecell}
\usepackage{amsmath,amssymb,amsfonts}
\usepackage{algorithm}
\usepackage{algorithmic}
\usepackage{graphicx}
\usepackage{textcomp}

\usepackage{microtype}
\usepackage{multirow}
\usepackage{booktabs}
\usepackage{rotating}
\usepackage{soul}
\usepackage{cancel}

\usepackage{url}

\usepackage[dvipsnames, table]{xcolor}
\usepackage{xcolor}
\usepackage{hyperref}

\usepackage{caption}
\journal{Medical Image Analysis}

\begin{document}

\verso{F. Tang \textit{et~al.}}

\begin{frontmatter}

\title{Hi-End-MAE: Hierarchical encoder-driven masked autoencoders are stronger vision learners for medical image segmentation}%

\author[1,2]{Fenghe \snm{Tang}}

\author[3]{Qingsong \snm{Yao}}
\author[1,2]{Wenxin \snm{Ma}}
\author[1,2]{Chenxu \snm{Wu}}
\author[1,2,4]{Zihang \snm{Jiang}\corref{cor1}}
\author[1,2,4,5]{S. Kevin \snm{Zhou}\corref{cor1}}

\cortext[cor1]{Corresponding authors: skevinzhou@ustc.edu.cn (S Kevin Zhou) and jzh0103@ustc.edu.cn (Z.H. Jiang).}
  
\address[1]{School of Biomedical Engineering, Division of Life Sciences and Medicine, University of Science and Technology of China (USTC), Hefei, Anhui, 230026, P.R. China}

\address[2]{Center for Medical Imaging, Robotics, and Analytic Computing \& LEarning (MIRACLE), Suzhou Institute for Advanced Research, USTC, Suzhou 215123, China
}
\address[3]{Stanford University, Palo Alto, California, 94305, United State}
\address[4] {State Key Laboratory of Precision and Intelligent Chemistry, USTC, Hefei, Anhui 230026, China}
\address[5] {Key Laboratory of Intelligent Information Processing of Chinese Academy of Sciences (CAS), Institute of Computing Technology, CAS, Beijing, 100190, China}

\begin{abstract}
Medical image segmentation remains a formidable challenge due to the label scarcity. Pre-training Vision Transformer (ViT) through masked image modeling (MIM) on large-scale unlabeled medical datasets presents a promising solution, providing both computational efficiency and model generalization for various downstream tasks. However, current ViT-based MIM pre-training frameworks predominantly emphasize local aggregation representations in output layers and fail to exploit the rich representations across different ViT layers that better capture fine-grained semantic information needed for more precise medical downstream tasks. To fill the above gap, we hereby present {\bf Hi}erarchical {\bf En}coder-{\bf d}riven {\bf MAE} ({\bf Hi-End-MAE}), a simple yet effective ViT-based pre-training solution, which centers on two key innovations: (1) Encoder-driven reconstruction, which encourages the encoder to learn more informative features to guide the reconstruction of masked patches; and (2) Hierarchical dense decoding, which implements a hierarchical decoding structure to capture rich representations across different layers. We pre-train Hi-End-MAE on a large-scale dataset of 10K CT scans and evaluated its performance across seven public medical image segmentation benchmarks. Extensive experiments demonstrate that Hi-End-MAE achieves superior transfer learning capabilities across various downstream tasks, revealing the potential of ViT in medical imaging applications. The code is available at: \url{https://github.com/FengheTan9/Hi-End-MAE}.

\end{abstract}


\end{frontmatter}

\section{Introduction}
\label{sec:introduction}

\begin{figure*}[tbp]
    \centering
    \includegraphics[width=0.99\linewidth]{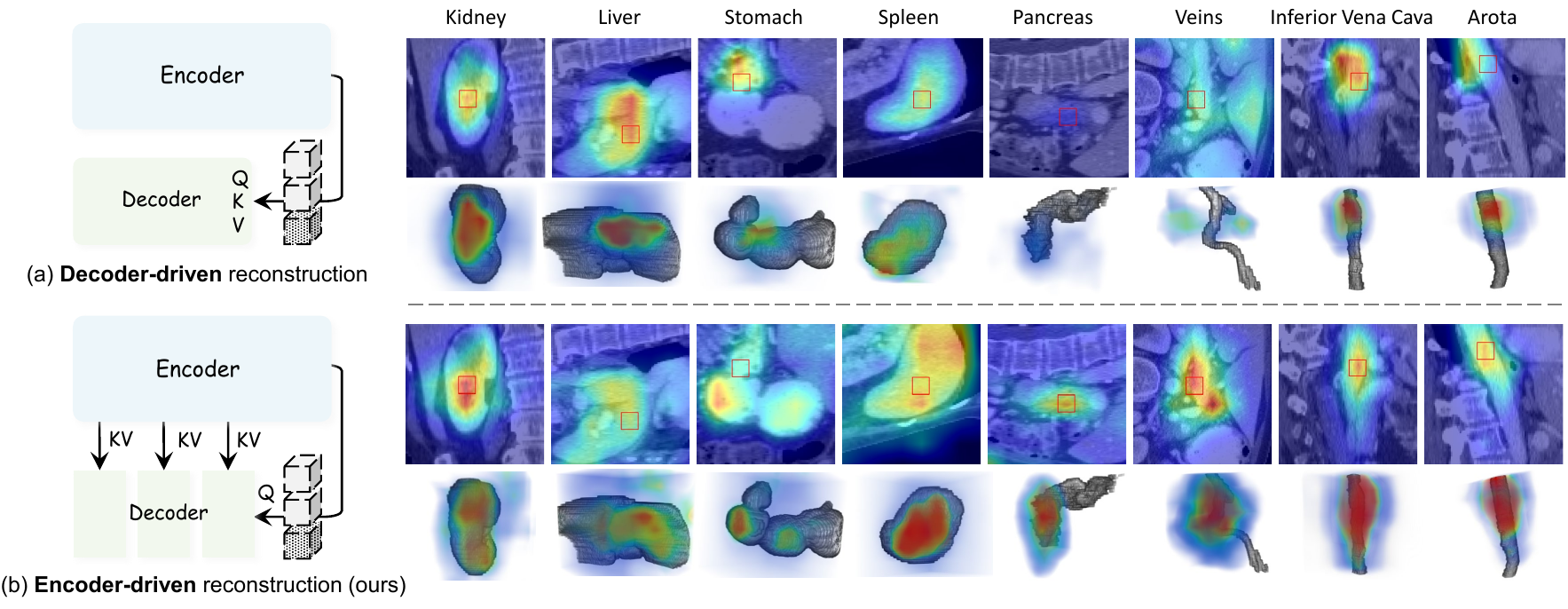}
    \vspace{-3mm}
    \caption{Decoder-driven vs. encoder-driven reconstruction. 
    Conventional MAE is based on (a) decoder-driven reconstruction and Hi-End-MAE is based on (b) encoder-driven reconstruction.    
    The slice-based (the first row) and volume-based (the second row) attention maps for query patches (red box) on different anatomical structures in the last layer of ViT, pre-trained by MAE and Hi-End-MAE, are visualized. The attention maps of MAE tend to attention on limited local contexts while those of Hi-End-MAE tend to be of more complete anatomical contexts, which are more instrumental to medical image segmentation.}
    \label{fig:vis_attention}
    \vspace{-4mm}
\end{figure*}


    

    

\begin{figure}[htbp]
    \centering
    \includegraphics[width=\linewidth]{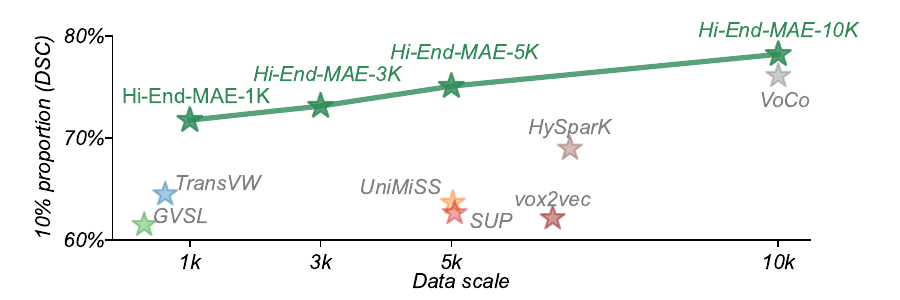}
    \vspace{-1mm}
    \includegraphics[width=\linewidth]{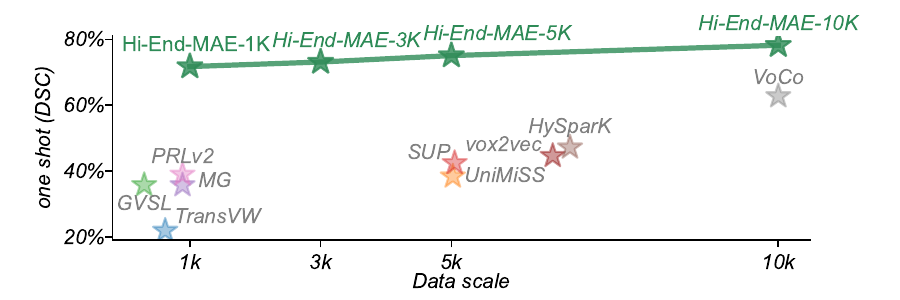}
    \vspace{-5mm}
    \caption{Performance comparisons against well-known medical self-supervised learning method using different pre-training data scales. Top figure represents the results fine-tuned by 10\% proportion of data, while the bottom figure represents fine-tuning with only one single 3D volume (one-shot).}
    \label{fig:vs}
    \vspace{-6mm}
\end{figure}

Deep learning demonstrates remarkable advancements in medical image analysis~\citep{deepmia}; however, it’s significantly hindered by the labor-intensive and time-consuming annotations by clinicians and experts~\citep{sslmia}. Especially in 3D medical image segmentation, the limited annotated data presents significant challenges for medical tasks~\citep{ssl3d, abd1k, word}. 
To alleviate this burden, a branch of self-supervised learning (SSL) methods are developed to pre-train the vision encoder on massive unlabeled data using proxy tasks~\citep{simclr,moco,dino,mae} and transfer it to downstream tasks. This paradigm presents a promising solution in label-efficient learning~\citep{mg,voco,hyspark}. 

Nevertheless, given the constraints of computational resources and increase in the amount of unlabeled medical data, pre-training on large-scale 3D medical datasets poses significant challenges. This limitation has driven the need for resource-efficient and performance-scaling pre-training frameworks. In this context, a representative Masked Image Modeling (MIM) technique, Masked Autoencoder (MAE)~\citep{mae}, has emerged as a promising solution, which pre-trains Vision Transformer (ViT)~\citep{vit} by handling only a small subset of visible patches. This computationally efficient approach offers significant advantages for pre-training on large-scale 3D medical datasets~\citep{mim}. However, the MAE framework imposes architectural constraints, requiring implementation through token-independent vanilla ViT architectures to boost pre-training efficiency. Compelling evidence from medical segmentation benchmarks reveals a fundamental limitation: segmentation networks built upon vanilla ViT backbones  exhibit inferior performance compared to convolutional neural network~\citep{unet, mednext} or hybrid network~\citep{hyspark}. This performance gap originates from ViT's intrinsic lack of spatial inductive biases —— a critical shortcoming that becomes acutely detrimental when trained from scratch with limited annotated datasets~\citep{unetr, swinunetr, unet, mednext, hyspark}. These factors lock the potential of ViT in medical image analysis.

The success of pre-trained Vision Transformer (ViT) in natural image processing~\citep{dino, mae} has motivated us to investigate its potential in medical vision tasks. Recent study reveal that masked image modeling (MIM) is inherently suited for low-level tasks as it effectively learns localized attention patterns that compensate for the limited inductive bias in ViT~\citep{lowlevel}. Crucially, while pre-trained ViTs exhibit remarkable few-shot learning capabilities in natural image~\citep{dino, mae, simmim, beit}, this generalization capacity remains underexplored in medical domains, which is essential for tackling segmentation tasks with label-scarcity.

Most advanced MIM methods designed for ViT, such as MAE, are tailored for natural images~\citep{mae, mae3d, cae, peco, simmim}, where separate encoder and decoder are employed for representation and reconstruction. We categorize this series of methods as {\bf decoder-driven reconstruction} (Fig.~\ref{fig:vis_attention} (a)). Although this approach partially liberates the encoder's representational capacity, the decoder still plays a crucial role in the reconstruction~\citep{cae}, thus it does not perform well enough in medical image pre-training. {As shown in Fig.~\ref{fig:vis_attention}, when visualizing the attention map at anatomies, we observe that MAE's local query attention demonstrates limited adaptability across diverse anatomical contexts (\textit{e.g.}, limited tubular and clustered attention patterns).}
We believe that this limitation likely stems from the MAE's architectural constraints in capturing hierarchical medical semantics; specifically, they fail to adequately leverage high-quality, rich representations across ViT's different layers during pre-training (shown in Fig.~\ref{fig:total_explain_er} and Fig.~\ref{fig:total_explain_sv}). Unlike conventional medical architectures (\textit{e.g.}, U-Net~\citep{unet} and its variants~\citep{mednext, unetr, swinunetr}) that systematically leverage multi-scale feature learning through dense skip-connections, ViT implementations under typical MIM frameworks tend to prioritize local aggregation representation in the output layer, which might miss the potential gain brought by different layer rich anatomical structure information gains (shown in Fig.~\ref{fig:attn_map}). 

Based on the above limitation, a natural insight arises: \textit{Is it possible to introduce efficient hierarchical local representation learning in MIM by solving proxy reconstruction tasks}? Different from the previous decoder-driven reconstruction, we try to introduce a simple yet effective \textbf{Hi}erarchical \textbf{En}coder-\textbf{d}riven dense-decoding architecture (Hi-End-MAE) to solve this problem (shown in Fig.~\ref{fig:vis_attention} (b)). There are two key innovative ideas behind Hi-End-MAE: \textit{\textbf{(i) Encoder-driven reconstruction}}:  utilizing decoder tokens to query visible encoded representations efficiently. The attention-weighted value further reconstructs the masked patches. 
This mechanism reduces the role of the decoder for reconstruction and directly establishes the relationship between representation quality and reconstruction ability, which makes the encoder learn stronger representation (encoder token values in Fig.~\ref{fig:total_explain_er} and Fig.~\ref{fig:total_explain_sv}). \textit{\textbf{(ii) Hierarchical dense decoding}}: performing densely bottom-up hierarchical decoding to learn more informative anatomical patterns between different layers. During bottom-up encoder-driven dense decoding, Hi-End-MAE progressively reduces the decoder’s workload, which compels the encoder to learn informative, hierarchical representations to compensate for decoding information loss.

Compared to other medical SSL methods, our Hi-End-MAE not only learns higher-quality local representations, compensating the inherent inductive bias limitations of ViT, but also captures localized anatomical patterns across different layers, which are crucial for medical imaging tasks and friendly for up-downstream alignment. Furthermore, by using visible encoder tokens for decoding, our Hi-End-MAE is faster and stronger than MAE, making it well-suited for large-scale datasets pre-training. Through an extensive empirical evaluation across seven downstream medical datasets, we demonstrate that:

\begin{itemize}
  \item Encoder-driven reconstruction in Hi-End-MAE could learn strong representations by solving proxy reconstruction tasks (Fig.~\ref{fig:framework}), which outperforms other well-known medical SSL methods in one-shot segmentation tasks across six medical datasets (Fig.~\ref{fig:vs} (top) and Table~\ref{Tab.oneshot}) and different proportion fine-tuning on three medical datasets (Fig.~\ref{fig:vs} (bottom) and Table~\ref{tab:diff_proportions}). 

  \item Benefiting from hierarchical dense decoding, Hi-End-MAE learns richer localized anatomical representations across different layers than MAE (Fig.~\ref{fig:total_explain_er} and Fig.~\ref{fig:total_explain_sv}). Token-query-based attention map visualization also reveals that Hi-End-MAE could learn strong local patterns on specific organs, such as tubular attention and clustered attention (Fig.~\ref{fig:vis_attention}).

  \item Hi-End-MAE is also generalizable and efficient. The robust local patterns learned by Hi-End-MAE can effectively generalize to other modalities, such as MRI (Table~\ref{tab:brats}). Additionally, thanks to encoder-driven reconstruction, Hi-End-MAE requires less computational cost than MAE (Table~\ref{tab:ablation1} and Table~\ref{tab:ablation2}). This reduction in computation is linearly related to the mask ratio, which greatly reduces the computational cost of large-scale 3D medical image pre-training while achieving powerful medical representations.
    
\end{itemize}


\section{Related Works}

\subsection{Masked image modeling}
Driven by BERT~\citep{beit}, masked image modeling (MIM) aims to remove or corrupt portions of the visual input and learn to predict the corrupted ones~\citep{inpaint, simmim, mae, spark, highlevel, cae, peco}. These approaches have been studied to reveal their ability to learn local attention patterns~\citep{lowlevel} and demonstrate better transferability to downstream tasks, such as segmentation and detection~\citep{mae, simmim, cmae, spark, hyspark}. The most representative of these methods is Masked Autoencoders (MAE)~\citep{mae}, which achieves efficient pre-training by dropping masked tokens. Although MAE uses an asymmetric design for reconstruction, the decoder still plays a significant role in reconstruction, limiting the quality of representation learning~\citep{cae}. Despite recent efforts~\citep{cae, peco} to address this issue, they overlook the importance of different layer informative anatomical representation learning in visual pre-training tasks~\citep{spark, hyspark} and struggle to balance both efficiency and representational capability. 

Given the importance of enhancing localized anatomical representation learning, the core idea of our approach is integrating hierarchical encoder-driven reconstruction into the decoding process, reducing the decoder's workload in reconstruction to compel the encoder to assume a greater role in the reconstruction task.

\begin{figure*}[ht!]
    \centering
    \includegraphics[width=0.95\linewidth]{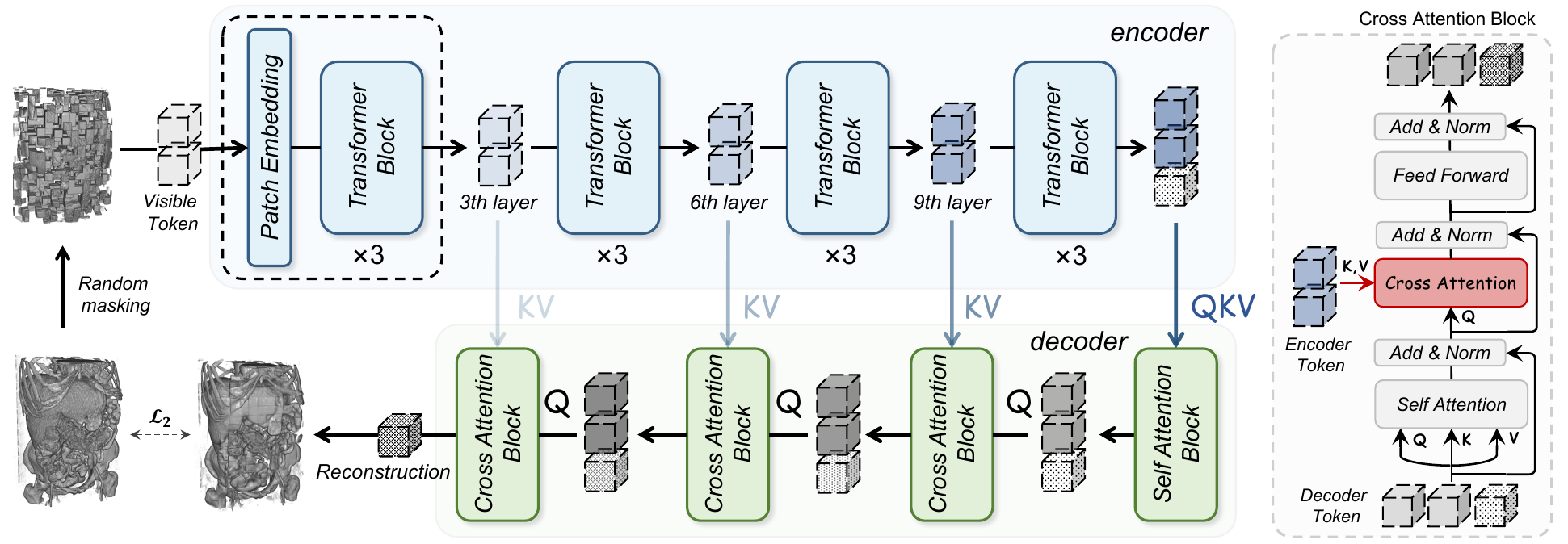}
    \vspace{-1mm}
    \caption{\textbf{The overall framework of Hi-End-MAE}. The Encoder-driven Dense Decoding architecture uses encoder representations to guide the decoder bottom-up dense reconstruction. The encoder (blue) is a Vision Transformer (ViT), which only processes the visible patches (blue cube). The decoder (green) incorporates a cross-attention mechanism, feeding in a full set of token \textit{i.e.} visible token (grey cube) and learnable masked token (mosaic cube) to query the encoder representation (blue arrow) for encoder-driven reconstruction.}
    \label{fig:framework}
    \vspace{-4mm}
\end{figure*}

\subsection{Self-supervised learning for medical imaging.}
Due to the scarcity of labeled medical images, self-supervised learning for medical images is a promising task~\citep{survey}. Existing medical SSL methods are mainly based on contrastive learning, conducted with strong data augmentation \textit{e.g.}, rotate~\citep{swinunetr, rotate} and multi-view crops~\citep{prlv2, unimiss, voco, gvsl, vox2vec}. However, most of these learn modality-specific high-level semantic representations~\citep{highlevel}, which introduce strong biases in downstream tasks with different data distributions~\citep{sslmia, bias}. In contrast, introducing MIM methods in medical image pre-training~\citep{mim, mae3d, hyspark} presents a promising avenue for addressing the above challenges~\citep{survey, mim}. However, most of these methods rarely prioritize the learning of anatomical semantics and downstream adaptation, both of which are crucial for medical visual tasks.

In this paper, we propose a simple yet effective MIM method. Unlike previous works, we emphasize the importance of enhancing the encoder's localized anatomical representation learning and downstream adaptability. By leveraging the encoder’s localized anatomical representations for dense decoding, our approach not only minimizes the role of the decoder in reconstruction, unleashing the potential of the encoder in medical visual learning, but also enables seamless adaptation of the encoder to downstream tasks.

\section{Methodology}

The overall framework of Hi-End-MAE is illustrated in Fig.~\ref{fig:framework}, which consists of two components: The \textbf{encoder} for localized anatomical representation across different layers and the \textbf{hierarchical encoder-driven dense decoder} for reconstruction.

\subsection{Masked encoding}
\noindent\textbf{Tokenify and masking.} We first tokenize the input 3D volume $x \in \mathbb{R}^{H \times W\/  \times D}$ into a sequence of $N$ volume patch tokens $\left \{  x_{i}\right \}_{i=1}^{N} x_{i} \in \mathbb{R}^{T \times P^{3}\/}$, where $P$ is patch size, $T = HWD / P^{3}$ is number of tokens. Following the MAE~\citep{mae}, we mask out a large ratio ($\gamma$) of patches and feed the visible patches $\left \{  x_{j}^{v}\right \}_{j=1}^{(1-\gamma) \cdot N}$ into the encoder.

\noindent\textbf{Masked encoding.} The encoder $\mathcal{F}$ maps the visible patches $x^{v}$ to multi-layer embedding features $ \left \{  Z_{l}^{v}\right \}_{l=1}^{L}$. We use ViT to form our encoder, which consists of $L$ Transformer layers. It first embeds the visible tokens $x^{v}$ by linear projection as token embeddings and adds the 3D positional embeddings followed by~\citep{mae3d}.

\subsection{Hierarchical encoder-driven dense decoding}
Unlike previous methods~\citep{mae, mae3d, cae}, our proposed Hi-End-MAE fully utilizes encoder representations to guide
the decoder for bottom-up dense reconstruction. The decoder $\mathcal{D}$ consists of a self-attention block and $B$ cross-attention blocks, which map the mask query token to the pixel space. We feed a full set of tokens $ X_{\mathcal{D}} = \left \{  x_{i}^{\mathcal{D}}\right \}_{i=1}^{N}$, consisting of last layer encoder tokens and learnable mask tokens with adding decoding position embeddings, into the decoder. Before encoder-driven reconstruction, we employ a few self-attention layers for early decoding adaption.

\noindent\textbf{Encoder-driven reconstruction.} To enable encoder-guided reconstruction which further enhances the encoder representation capacity, we introduce the cross-attention mechanism into the decoding process. Specifically, the decoding tokens ($Q_{\mathcal{D}}$) query visible encoder representations $(V_{\mathcal{F}})$ (blue arrow in Fig.~\ref{fig:framework}), and then the queried values ($O_{\mathcal{D}}$) (grey and mosaic cube in Fig.~\ref{fig:framework}) are used for the next stage decoding. Under encoder-driven reconstruction, the reconstruction quality directly depends on the encoder representation quality ($V$). This operation compels the encoder to learn stronger representations to compensate for the loss of reconstruction information. It can be represented as:

\begin{equation}
    \text{Attention} = \text{Softmax}(\frac{QK ^{T}}{\sqrt{d_{k}}})V
\end{equation}

\begin{equation}
    \check{X}_{\mathcal{D}} = \text{Attention}(Q_{\mathcal{D}}, K_{\mathcal{D}}, V_{\mathcal{D}}) + X_{\mathcal{D}}
    \label{equ:self}
\end{equation}

\begin{equation}
    X_{\mathcal{F}2\mathcal{D}} = \text{Attention}(\check{Q}_{\mathcal{D}}, K_{\mathcal{F}}, V_{\mathcal{F}}) + \check{X}_{\mathcal{D}}
    \label{equ:cross}
\end{equation}

\begin{equation}
    O_{\mathcal{D}} = \text{FeedForward}(X_{\mathcal{F}2\mathcal{D}}) + X_{\mathcal{F}2\mathcal{D}}
\end{equation}

\noindent where the Eqs.~(\ref{equ:self}) and (\ref{equ:cross}) represent self-attention and cross-attention, respectively. $Q_{\mathcal{D}}$, $K_{\mathcal{D}}$, $V_{\mathcal{D}}$ and $\check{Q}_{\mathcal{D}}$ are linear projection from $X_{\mathcal{D}}$ and $\check{X}_{\mathcal{D}}$, respectively. Key ($K_{\mathcal{F}}$) and value ($V_{\mathcal{F}}$) in Equ.\ref{equ:cross} are linear projection from $Z^{v}$. $O_{D}$ is the output queries. $d_{k}$ is the feature dimension of $K$.

\noindent\textbf{Hierarchical dense decoding.} Considering the importance of localized anatomical representations across different layer for medical segmentation tasks, we introduce hierarchical bottom-up dense decoding. Specifically, in the bottom-up decoding process, the decoding at different stages queries the corresponding encoding representation (blue arrow in Fig.~\ref{fig:framework}). It can be represented as:

\begin{equation}
    X_{\mathcal{F}2\mathcal{D}}^{b,l} = \text{Attention}(\check{Q}_{\mathcal{D}}^{b}, K_{\mathcal{F}}^{l}, V_{\mathcal{F}}^{l}) + \check{X}_{\mathcal{D}}^{b}
    \label{equ:cross2}
\end{equation}

\noindent where the $\check{X}_{\mathcal{D}}^{b},  b\in \left \{ 1,2,..,B \right \} $ is the $b$-th stage decoding token. $\check{Q}_{\mathcal{D}}^{b}$ is query token linear projected from $\check{X}_{\mathcal{D}}^{b}$. $K_{\mathcal{F}}^{l}$ and $V_{\mathcal{F}}^{l}$ are linear projected by the correspond $l$-th layer encoder feature. Given that dense decoding is a bottom-up decreasing information process, the closer to the output, the less representation information is supported by the encoder feature values (lighter arrow colors indicating weaker representations in Fig.~\ref{fig:framework}), which compels the encoder to learn stronger and richer localized anatomical representations.

\noindent\textbf{Reconstruction and loss.} Followed by MAE~\citep{mae, mae3d}, we use a linear projection layer for the final reconstruction and optimize a mean square error loss ($\mathcal{L}_{2}$) for masked region reconstruction.

\noindent\textbf{Quantitative evaluation of encoder representation.} Followed previous works~\citep{svd, er2, er3}, we use singular values and effective rank~\citep{effectiverank} to quantify our encoder representation ability.

Given the matirx $A  \in \mathbb{R}^{m \times n}$ and its singular values $\left \{ \sigma_{i}  \right \} _{i=1}^{min(m,n)} $, the effective rank $\rho(A)$ is defined as:

\begin{equation}
    \rho(A) = -\sum_{i=1}^{min(m,n)}\bar{\sigma_{i}}log(\bar{\sigma_{i}})
\end{equation}

\noindent where $\bar{\sigma_{i}}= \sigma_{i}/\sum_{k} \sigma_{i}$ is $i$-th normalized singular value. In this paper, matrix $A$ represents the attention values ($V$) of each encoder layer.

\noindent\textbf{Computational complexity of encoder-driven reconstruction.} Conventional decoder-driven reconstruction methods perform self-attention over the full set of $N$ tokens, resulting in quadratic complexity $\mathcal{O}(N^2 d_{k})$, where $N$ is the total number of tokens and $d_{k}$ denotes the feature dimension. On the contrary, our proposed decoder-driven reconstruction could further reduce complexity by querying only visible encoded tokens. Specifically, only visible encoder tokens as Key $(K_{\mathcal{F}})$ and Value $(V_{\mathcal{F}})$ are participated in calculation, with the preserved token count $M = N \cdot (1-\gamma) $, where $\gamma \in (0,1)$ is the masking ratio. This reduces the complexity of computing the attention similarity matrix from $\mathcal{O}(N^2 d_{k})$ to $\mathcal{O}(N M d_{k}) = \mathcal{O}(N^2 (1-\gamma) d_{k})$. For an encoder-driven reconstruction layer, the total complexity is given as:
\begin{equation}
    O_{\text{total}} = \mathcal{O}(N^2 (1-\gamma) d_{k}) + \mathcal{O}(N d_{k}^2).\label{eq:comp} 
\end{equation}
Note that the first term in Eq. (\ref{eq:comp}) is dominant. In our experiment, we set $\gamma=0.75$, which means significant saving in computation in practice.


\section{Experiments}

\subsection{Datasets}


\noindent\textbf{Pre-training datasets.} As shown in Table~\ref{tab:dataset}, a total of 13 public CT datasets, consisting of 9995 ($\approx$10k) CT scans, are curated to form our pre-training dataset, including BTCV~\citep{btcv}, Sliver07~\citep{sliver07}, CT-ORG~\citep{ctorg}, FLARE'22~\citep{flare}, CHAOS~\citep{chaos}, NaH-Seg~\citep{han}, KiPA22~\citep{kipa1, kipa2, kipa3, kipa4}, COVID-19~\citep{convid}, Pancreas-CT~\citep{pan}, LiTS~\citep{lits}, AbdomenCT-1k~\citep{abd1k}, LUNA16~\citep{luna}, and AbdomenAtlasMini 1.0~\citep{atlas10}. Existing annotations or labels are not utilized from these datasets during pre-training. The pre-train datasets are interpolated to the isotropic voxel spacing of $1.5\ mm$. Intensities are scaled to $[-175, 250]$, then normalized to $[0, 1]$. We crop sub-volumes of $96\times96\times96$ voxels as input. Details are provided in the Appendix.

\begin{table}[!t]
\caption{Overview of pre-train and downstream dataset.\label{tab:dataset}}
\vspace{-2mm}
\centering
\resizebox{0.9\linewidth}{!}
{
\begin{tabular}{lr c c}
\Xhline{1px} 
Dataset & \# of volumes & Pre-train & Downstream\\
\hline
BTCV & 50 & \checkmark & \checkmark \\ Sliver07 & 20 & \checkmark & \checkmark \\
CT-ORG & 140 & \checkmark & \checkmark \\ FLARE'22 & 2300 & \checkmark & \checkmark \\
CHAOS & 40 & \checkmark & \\ 
HaN-Seg & 42 & \checkmark &  \\
KiPA22 & 70 & \checkmark & \\
COVID-19 & 10 & \checkmark &  \\
Pancreas-CT & 82 & \checkmark & \\
LiTS & 134 & \checkmark &  \\
AbdomenCT-1k & 1062 & \checkmark & \\
LUNA16 & 888 & \checkmark &  \\
AbdomenAtlas 1.0 & 5195 & \checkmark & \\
WORD & 150  &  & \checkmark \\
AMOS & 600 & & \checkmark \\
BraTS 21 & 1200 &  & \checkmark \\

\Xhline{1px}
\end{tabular}
}
\vspace{-4mm}
\end{table}

\noindent\textbf{Downstream datasets.} To evaluate the effectiveness of our method, we conduct downstream experiments on seven public datasets for medical image segmentation~\citep{lay2013rapid}, \textit{e.g.}, BTCV~\citep{btcv}, CT-ORG~\citep{ctorg}, Sliver07~\citep{sliver07}, WORD~\citep{word}, AMOS~\citep{amos}, FLARE'22~\citep{flare} and BraTS21~\citep{brats21}. To better assess the representation capacity of the pre-trained model, we employ the first six datasets for one-shot segmentation tasks. Additionally, following previous works~\citep{ratio1, ratio2, ratio3, ratio4}, we selected AMOS~\citep{amos}, FLARE'22~\citep{flare} and BTCV~\citep{btcv} datasets with official training-validation split for downstream experiments with 1\%, 10\% and 100\% data proportions. To assess the model's cross-modality generalization capability, we transfer the pre-trained model from the CT domain to the MRI (i.e. adapt in BraTS  
 21~\citep{brats21}) for further evaluation. We adopt consistent settings as previous works~\citep{unetr, swinunetr, voco}. More pre-training details are provided in the Appendix.

\renewcommand{\multirowsetup}{\centering}  
\begin{table*}[!]
\centering
\caption{Comparison of different methods for one-shot segmentation on BTCV, CT-ORG, Sliver07, WORD, AMOS and FLARE'22. \textbf{val} (bold) / \underline{val} (underline) : top method / second method. $\dagger$ denotes we utilize official pre-training weights.}
\vspace{-2mm}
\resizebox{0.95\linewidth}{!}
{
\begin{tabular}{l l | l | cccccc | c}
\Xhline{1px} 
\multicolumn{2}{c|}{Pretrain Method} & \multirow{2}{*}{\# Volume} & \multicolumn{6}{c|}{ Dataset (DSC \%)}  & \multirow{2}{*}{Avg} \\
\cline{1-2}\cline{4-9}
Method & Network & & BTCV & CT-ORG & Sliver07 & WORD & AMOS & FLARE'22 & \\
\hline
\multicolumn{2}{c|}{\textit{{\color{Gray} Training from scratch}}}  & & & & & & & & \\
- & UNETR  & WACV'22 & 24.27 & 49.08 & 80.87 & 30.89 & 10.06 & 26.30 & 36.91 \\
- & SwinUNETR & CVPR'22 & 27.71 & 55.28 & 80.91 & 43.68 & 9.59 & 35.89 & 42.17 \\
\hline
\multicolumn{2}{c|}{\textit{{\color{Gray} General self-supervised methods}}} & & & & & & & & \\
SparK & MedNeXt & ICLR'23 & 30.69 & 62.24 & 84.89 & 50.64 & 13.34 & 36.48 & 46.37 \\
MAE & UNETR  & CVPR'22 & 62.04 & \underline{69.75} & 78.02 & \underline{69.06} & 38.05 & \underline{62.35} & \underline{63.21} \\
\hline
\multicolumn{2}{c|}{\textit{{\color{Gray} Medical self-supervised methods}}} & & & & & & & & \\
MG$^{\dagger}$ & 3D U-Net & MICCAI’20 & 29.27 & 51.12 & 67.40 & 27.95 & 11.67 & 27.30 & 35.78 \\
TransVW$^{\dagger}$ & 3D U-Net & TMI’21 & 5.63 & 34.74 & 75.77 & 7.23 & 3.66 & 4.81 & 21.97 \\
UniMiSS$^{\dagger}$ & MiT & ECCV’22 & 32.95 & 60.24 & 75.96 & 23.03 & 13.46 & 24.92 & 38.42 \\
SUP$^{\dagger}$ & SwinUNETR & CVPR'22 & 28.75 & 56.72 & 78.99 & 46.95 & 9.94 & 33.72 & 42.51 \\
PRLv2$^{\dagger}$ & 3D U-Net & TPAMI'23 & 24.01 & 55.86 & 83.35 & 31.69 & 11.54 & 27.71 & 39.02 \\
GVSL$^{\dagger}$ & 3D U-Net & CVPR'23 & 24.86 & 54.57 & 60.53 & 37.87 & 10.84 & 26.33 & 35.83 \\
vox2vec$^{\dagger}$ & 3D U-Net & MICCAI'23 & 35.29 & 62.91 & 72.94 & 49.37 & 13.44 & 34.11 & 44.67 \\
HySparK$^{\dagger}$ & MedNeXt + ViT & MICCAI'24 & 35.81 & 60.83 & 80.61 & 53.27 & 15.31 & 37.54 & 47.22  \\
VoCo$^{\dagger}$ & SwinUNETR & CVPR'24 & \underline{63.33} & 65.12 & \underline{87.43} & 64.24 & \underline{38.80} & 57.66 & 62.76 \\
\rowcolor{gray!15} Hi-End-MAE & UNETR & ours & \textbf{69.59} & \textbf{71.09} & \textbf{91.88} & \textbf{73.52} & \textbf{46.21} & \textbf{63.22} & \textbf{69.25} \\
\Xhline{1px} 

\end{tabular}
}
\label{Tab.oneshot}
\end{table*}

\begin{figure*}[t]
    \centering
    \includegraphics[width=0.96\textwidth]{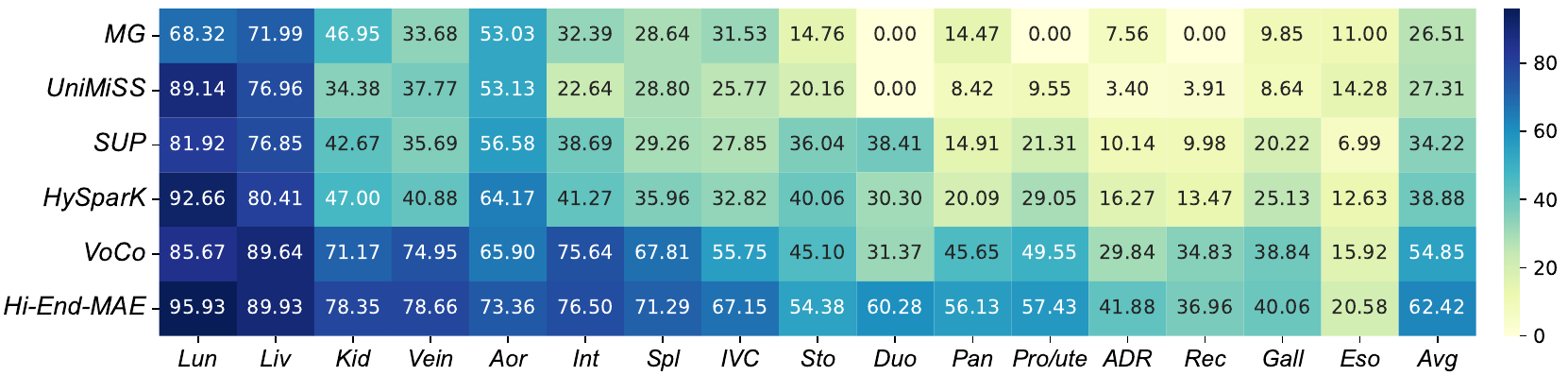}
    \vspace{-2mm}
    \caption{Comparative analysis one-shot segmentation results across 16 targets in terms of DSC (\%) performance. The abbreviations Lun, Liv, Kid, Vein, Aor, Int, Spl, IVC, Sto, Duo, Pan, Pro/ute, ADR, Rec, Gall and Eso correspond to Lung, Liver, Kidney, Veins, Aorta, Intestine, Spleen, Inferior Vena Cava, Stomach, Duodenum, Pancreas, Prostate/Uterus, Adrenal Gland, Rectum, Gallbladder and Esophagus, respectively.}
    \label{fig:oneshotlist}
    \vspace{-4mm}
\end{figure*}

\subsection{Implementation details}

\noindent\textbf{Experiment settings.} 
Following previous works~\citep{mae3d, benchmark, hyspark}, we adopt UNETR~\citep{unetr} as the downstream networks. To achieve upstream-downstream alignment, we set three decoding stages, corresponding to the encoder representations at layers 3-th, 6-th, and 9-th, \textit{e.g.} same skip-connection layers in UNETR, to guide decoder reconstruction. For pre-training tasks, we train with the AdamW optimizer, an initial learning rate of 1e-4, and a cosine-annealing scheduler for all experiments. The pre-training experiments use a batch size of 192 and train the model for 400K steps. For fair downstream comparisons, detailed training hyper-parameters settings for fine-tuning and inference are the same as previous works~\citep{unetr, swinunetr, voco, hyspark}. Details are provided in the Appendix.

\noindent\textbf{Comparison methods.} We select both general and medical self-supervised methods for a comprehensive comparison. First, we pre-train and compare with the well-known MIM method MAE~\citep{mae, mae3d} and SparK~\citep{spark} with the same experiment settings. In addition, we choose nine advanced and well-known medical self-supervised methods: Models Genesis (MG)~\citep{mg}, TransVW~\citep{transvw}, UniMiSS~\citep{unimiss}, Swin UNETR Pre-trained method (SUP)~\citep{swinunetr}, PRLv2~\citep{prlv2}, GVSL~\citep{gvsl}, vox2vec~\citep{vox2vec}, HySparK~\citep{hyspark} and VoCo\footnote{official 10K ct-scan pre-training weights}~\citep{voco}. To ensure a fair comparison, we load the official pre-trained weights for all medical SSL methods before fine-tuning.




\subsection{Experiments on downstream tasks}
\noindent\textbf{Overall one-shot medical segmentation.} We first conduct one-shot segmentation on six datasets, as shown in Table~\ref{Tab.oneshot}. Our method substantially outperforms both general and medical self-supervised approaches, achieving a DSC score of 69.25\%, at least 6.04\% higher than all compared methods. The Hi-End-MAE pre-training yields a 32.34\% improvement over training from scratch. Compared to the recent contrastive-based method VoCo~\citep{voco}, which uses similarly-sized pre-training datasets (10k), Hi-End-MAE shows consistent improvements of 6.26\%, 5.97\%, 4.45\%, and 5.56\% on BTCV~\citep{btcv}, CT-ORG~\citep{ctorg}, Sliver07~\citep{sliver07}, and FLARE'22~\citep{flare}, respectively. Notably, on datasets unseen during pre-training, \textit{e.g.}, WORD~\citep{word} and AMOS~\citep{amos}, our method demonstrates promising performance, higher than VoCo with DSC of 9.28\% and 7.41\%, respectively. Additionally, it can be observed that apart from our method, which achieves a new state-of-the-art (SOTA) of 69.25\%, another ViT pre-trained with MIM also demonstrates notable generalization benefits, with MAE achieving 63.21\%.

\renewcommand{\multirowsetup}{\centering}  
\begin{table*}[t]
\centering
\caption{Comparison of different methods with different proportions on AMOS~\citep{amos}, FLARE'22~\citep{flare} and BTCV~\citep{btcv}. We report the DSC (\%) performance. \textbf{val} (bold) / \underline{val} (underline) : top method / second method. $\dagger$ denotes we utilize official pre-training weights. $\ddagger$ denotes the results are copied from~\citep{voco}.}
\vspace{-2mm}
\resizebox{0.99\linewidth}{!}
{
\begin{tabular}{l | l | cccc | cccc | cccc | c}
\Xhline{1px} 
\multirow{2}{*}{Pretrain Method} & \multirow{2}{*}{\# Volume} & \multicolumn{4}{c|}{AMOS} & \multicolumn{4}{c|}{FLARE'22} & \multicolumn{4}{c|}{BTCV}  & \multirow{2}{*}{Avg} \\
\cline{3-14}
& & 1\% & 10\% & 100\% & Avg & 1\% & 10\% & 100\% & Avg & 1\% & 10\% & 100\%$^{\ddagger}$ & Avg & \\
\hline
\multicolumn{2}{c|}{\textit{{\color{Gray} Training from scratch}}}  & & & & & & & & & & & & & \\
UNETR & WACV'22 & 23.67 & 60.06 & 77.02 & 57.45 & 22.47 & 56.46 & 70.81 & 49.91 & 28.05 & 42.85 & 79.82 & 50.24 & 52.53 \\
SwinUNETR & CVPR'22 & 28.94 & 63.45 & 82.51 & 58.30 &  35.89 & 63.38 & 75.38 & 58.21 & 27.71 & 51.33 &  80.53 & 53.19 & 56.56 \\
\hline
\multicolumn{2}{l|}{\textit{{\color{Gray} General self-supervised methods}}} & & & & & & & & & & & & & \\
SparK & ICLR'23 & 36.14 & 71.68 & 84.07 & 63.96 & 36.48 & 71.74 & 80.67 & 62.96 & 30.69 & 51.26 & - & - & -  \\
MAE & CVPR'22 & 54.67 & 72.94 & 83.61  & 70.40 & \underline{62.35} & 77.01 & 82.56 & \underline{73.97} & 62.04 & 75.01 & - & - & -   \\
\hline
\multicolumn{2}{l|}{\textit{{\color{Gray} Medical self-supervised methods}}} & & & & & & & & & & & & & \\
MG$^{\dagger}$ & MICCAI’20 & 25.72 & 46.94 & 62.99 & 45.21 & 27.30 & 48.18 & 57.33 & 44.27 & 29.27 & 38.04 &  81.45 & 49.58 & 56.97  \\
TransVW$^{\dagger}$ & TMI’21 & 18.72 & 66.91 & 82.58 & 56.06 & 4.81 & 62.07 & 75.78 & 47.55 & 5.63 & 8.42 & - & - & -   \\
UniMiSS$^{\dagger}$ & ECCV’22 & 29.49 & 66.34 & 79.92 & 58.58 & 24.92 & 60.99 & 74.71 & 53.54 & 32.95 & 47.08 & - & -  & -  \\ 
SUP$^{\dagger}$ & CVPR’22 & 25.60 & 64.95 & 82.45 & 57.66 & 33.72 & 60.35 & 74.96 & 56.34 & 28.75 & 49.67 & 81.54 & 53.32 & 55.77  \\ 
PRLv2$^{\dagger}$ & TPAMI’23 & 21.07 & 39.07 & 54.14 & 38.09 & 27.71 & 42.97 & 54.29 & 41.65 & 24.01 & 30.48 & 81.74 & 45.41 & 41.71  \\ 
GVSL$^{\dagger}$ & CVPR’23 & 24.25 & 63.45 & 81.38 & 56.35 & 26.33 & 59.54 & 73.27 & 53.04 & 24.86 & 41.79 & 81.87 & 49.50 & 52.96 \\
vox2vec$^{\dagger}$ & MICCAI’23 & 32.76 & 62.30 & 74.78 & 56.61 & 34.11 & 61.99 & 70.33 & 55.47 & 35.29 & 51.77 & - & - & -   \\
HySparK$^{\dagger}$ & MICCAI'24 & 34.50 & 64.32 & \textbf{85.58} & 61.46 & 37.54 & 73.60 & 82.35 & 64.49 & 35.81 & 51.54 & - & - & -  \\
VoCo$^{\dagger}$ & CVPR’24 & \underline{55.81} & \underline{73.34} & 84.44 & \underline{71.19} & 57.66 & \underline{78.84} & \underline{83.12} & 73.20 & \underline{63.33} & \underline{77.85} & \underline{83.85} & \underline{75.01} & 73.13 \\
\rowcolor{gray!15} Hi-End-MAE & ours & \textbf{60.35} & \textbf{75.84} & \underline{84.98} & \textbf{73.72} & \textbf{63.22} & \textbf{80.58} & \textbf{84.20} & \textbf{76.00} & \textbf{69.59} & \textbf{78.56} & \textbf{84.53} & \textbf{77.56} & \textbf{75.72} \\

\Xhline{1px}
\end{tabular}
}

\vspace{-3mm}
\label{tab:diff_proportions}
\end{table*}

\renewcommand{\multirowsetup}{\centering}  
\begin{table}[!]
\caption{Experimental one-shot results on BRATS 21~\citep{brats21}. TC, WT, and ET denote the tumor core, whole tumor, and enhancing tumor, respectively. \textbf{val} (bold) / \underline{val} (underline) : top method / second method. $\dagger$ denotes we utilize official pre-training weights.}
\vspace{-2mm}
\resizebox{1\linewidth}{!}
{
\begin{tabular}{ l | l | ccc | c}
\Xhline{1px} 
\multirow{1}{*}{Pretrain Method} & \multirow{1}{*}{\# Backbone} & TC & WT & ET  & \multirow{1}{*}{Avg} \\
\hline
\multicolumn{2}{c|}{\textit{{\color{Gray} Training from scratch}}}  & & & & \\
- & UNETR & 49.72 & 54.82 & 56.27 & 53.60 \\
- & SwinUNETR & 50.44 & 57.81 & \underline{59.60} & 55.95 \\
\hline
\multicolumn{2}{c|}{\textit{{\color{Gray} General self-supervised methods}}} & & & \\
MAE & UNETR & 34.54 & 45.51 & 39.24 & 39.76 \\
SparK & MedNeXt & 44.04 & 65.52 & 44.90 & 51.49 \\
\hline
\multicolumn{2}{c|}{\textit{{\color{Gray} Medical self-supervised methods}}} & & & & \\
MG$^{\dagger}$ & 3D U-Net & 28.25 & 34.08 & 33.52 & 31.95 \\
UniMiSS$^{\dagger}$  & MiT & 6.20 & 17.56 & 23.94 & 15.90 \\
SUP$^{\dagger}$ & SwinUNETR & 49.96 & 53.51 & 56.65 & 53.37  \\
GVSL$^{\dagger}$ & 3D U-Net & \underline{54.11} & \textbf{73.04} & 53.45 & \underline{60.20} \\
vox2vec$^{\dagger}$ & 3D U-Net & 23.93 & 33.33 & 21.97 & 26.41 \\
HySparK$^{\dagger}$ & MedNeXt + ViT & 40.05 & 44.36 & 46.98 & 43.79 \\
VoCo$^{\dagger}$ & SwinUNETR & 40.76 & 53.12 & 55.41 & 49.76  \\
\rowcolor{gray!15}  Hi-End-MAE & UNETR & \textbf{55.35} & \underline{68.25} & \textbf{67.46} & \textbf{61.45} \\

\Xhline{1px} 

\end{tabular}
}

\vspace{-6mm}
\label{tab:brats}
\end{table}

\begin{figure*}[!]
    \centering
    \vspace{2mm}
    \includegraphics[width=0.99\linewidth]{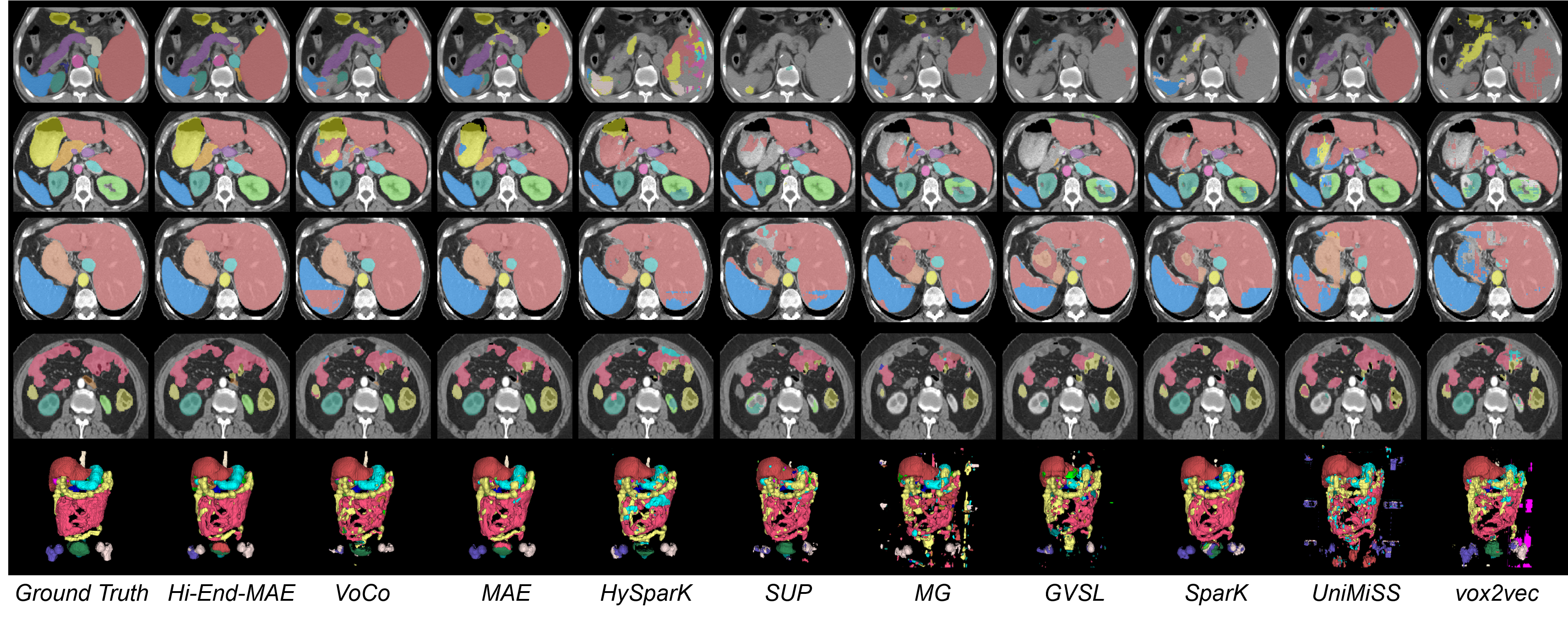}
    \vspace{-3mm}
    \caption{Qualitative visualization of one-shot segmentation results for AMOS~\citep{amos} (row 1), BTCV~\citep{btcv} (row 2), FLARE'22~\citep{flare} (row 3) and WORD~\citep{word} (row 4 and 5).}
    \vspace{-5mm}
    \label{fig:result}
\end{figure*}

\noindent\textbf{Target-specific analysis and visualization.} To comprehensively evaluate the advantages of Hi-End-MAE across different segmentation targets, we investigate the one-shot segmentation semantics results on six datasets. Comparative results are partially shown in Fig.\ref{fig:oneshotlist}. Hi-End-MAE achieves optimal results for most organs. Compared to the recent SOTA method VoCo~\citep{voco}, Hi-End-MAE shows significant improvements across organs of different scales, \textit{e.g.}, with enhancements of 28.91\%, 10.48\%, 10.26\%, and 11.40\% for the duodenum, pancreas, lung, and inferior vena cava, respectively. This indicates that Hi-End-MAE encourages the encoder to learn strong and rich localized anatomical representations. Additionally, we visualize segmentation results in Fig.\ref{fig:result}. Benefiting from high-quality localized anatomical representations,  Hi-End-MAE achieves more complete and accurate segmentation results.

\noindent\textbf{Comparison across various data proportions.} Following previous works~\citep{ratio1, ratio2}, we fine-tune pre-trained models on AMOS~\citep{amos}, FLARE'22~\citep{flare}, and BTCV~\citep{btcv} using 1\%, 10\%, and 100\% of the dataset, as shown in the Table~\ref{tab:diff_proportions}. Hi-End-MAE achieves an average DSC score of 75.72\% across various data proportions, clearly outperforming existing methods. Notably, our method surpasses the highest-performing compared method, VoCo, by 2.59\%. In terms of the gains from pre-training, our backbone network UNETR achieves a substantial improvement of 23.19\% compared to training from scratch(\textit{e.g.}, 52.53\% to 75.72\%), whereas VoCo's backbone SwinUNETR shows a gain of 16.57\% compared to training from scratch(\textit{e.g.}, 56.56\% to 73.13\%). These results prove that our improvement is consistent across different dataset proportions.

\begin{figure*}[t]
    \centering
    \includegraphics[width=0.92\textwidth]{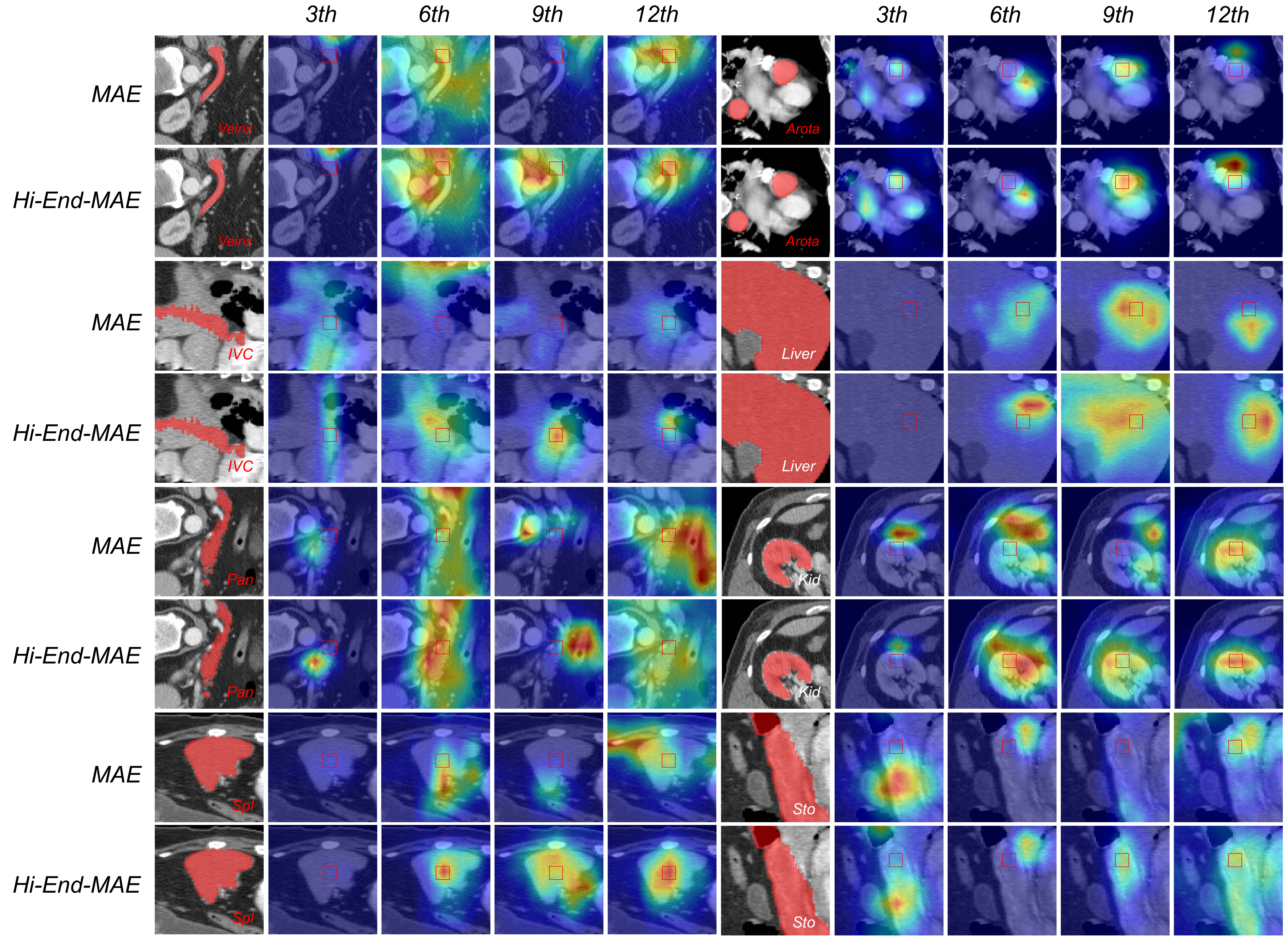}
    \vspace{-2mm}
    \caption{Visualization of attention maps in the 3th, 6th, 9th, and 12th layers of ViT-B/12$^{(1536)}$ for query patches (red box) on different organs, pre-trained by MAE and Hi-End-MAE. The attention maps correspond to the same attention head in both MAE and Hi-End-MAE encoder.}
    \label{fig:attn_map}
    \vspace{-2mm}
\end{figure*}

\begin{figure*}[!]
    \centering
    \includegraphics[width=0.92\textwidth]{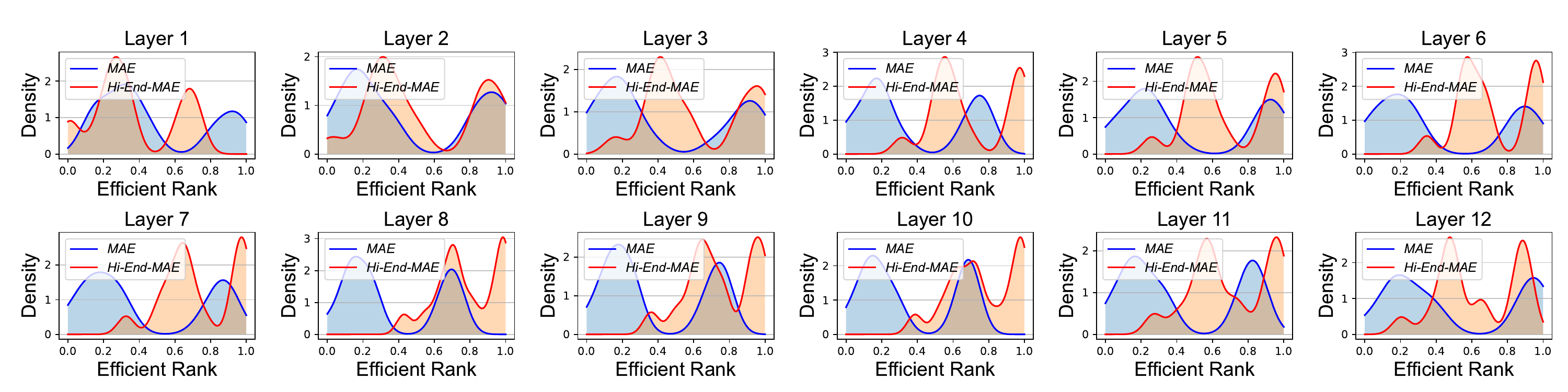}
    \vspace{-2mm}
    \caption{Comparison effective rank~\citep{effectiverank} distribution of attention values ($V$) with MAE and Hi-End-MAE in ViT-B/12$^{(1536)}$. A rightward shift indicates a richer effective rank, reflecting more diverse data representations and a more uniform distribution. This enables the model to capture greater information from the feature space.}
    \vspace{-4mm}
    \label{fig:total_explain_er}
\end{figure*}

\begin{figure*}[!]
    \centering
    \includegraphics[width=0.92\textwidth]{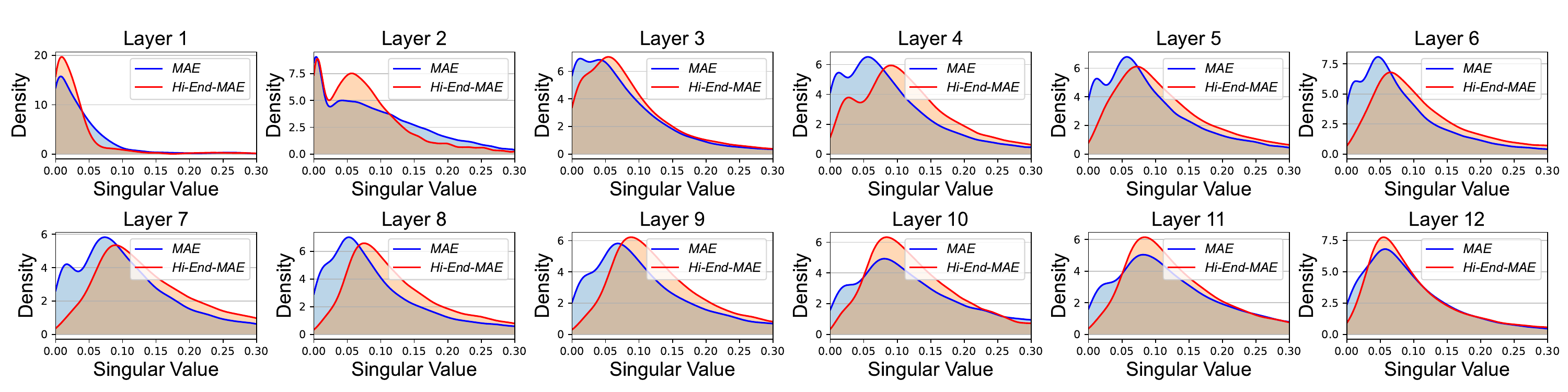}
    \vspace{-2mm}
    \caption{Comparison singular values distribution of attention values ($V$) with MAE and Hi-End-MAE in ViT-B/12$^{(1536)}$. A rightward shift indicates higher variance, suggesting stronger generalization ability and more diversified data representation.}
    \label{fig:total_explain_sv}
\end{figure*}

\noindent\textbf{Generalization to MRI modalities.} Our results prove that the performance gain of Hi-End-MAE is generalizable to other modalities. As shown in Table~\ref{tab:brats}, we evaluate the downstream on a widely used MRI dataset, \textit{e.g.} BraTS 21~\citep{brats21}. Hi-End-MAE achieves the highest DSC score, \textit{e.g.} 61.45\%, in one-shot segmentation, surpassing existing SOTA methods. Despite the significant differences in high-level semantics between MRI and CT, this generalization ability highlights that Hi-End-MAE effectively learns low-level representations that are independent of high-level semantics, thereby improving its performance across different modalities.

\renewcommand{\multirowsetup}{\centering}  
\begin{table}[!]
\caption{The effect of mask ratios.}
\vspace{-2mm}
\resizebox{1\linewidth}{!}
{
\begin{tabular}{ c | c c c c c c | c}
\Xhline{1px} 
\multirow{1}{*}{Mask Ratio} & \multirow{1}{*}{BTCV} & \multirow{1}{*}{CT-ORG}  & \multirow{1}{*}{Sliver07} & \multirow{1}{*}{WORD} & \multirow{1}{*}{AMOS} & \multirow{1}{*}{FLARE'22}  & \multirow{1}{*}{Avg} \\
\hline
75\% & 69.59 & 71.09 & 91.88 & 73.52 & 46.21 & 63.22 & 69.25 \\
80\% & 67.52 & 71.68 & 91.25 & 72.86 & 46.10 & 59.39 & 68.13 \\
85\% & 66.43 & 69.04 & 87.81 & 70.94 & 41.32 & 61.89 & 66.23 \\
90\% & 64.38 & 69.58 & 85.61 & 71.25 & 41.41 & 62.12 & 65.72 \\
95\% & 64.63 & 70.67 & 83.14 & 67.35 & 32.19 & 61.72 & 63.28 \\

\Xhline{1px} 

\end{tabular}
}
 \vspace{-6mm}
\label{tab:maskratio}
\end{table}

\renewcommand{\multirowsetup}{\centering}  
\begin{table*}[t]
\caption{Evaluation of encoder-driven reconstruction against decoder-driven reconstruction. We report the DSC score (\%) on three datasets with various proportions.  \textbf{val} (bold): top method.}
\vspace{-2mm}
\resizebox{1\linewidth}{!}
{
\begin{tabular}{l  | l | c | cccc | cccc | cccc | c}
\Xhline{1px} 
\multirow{2}{*}{Method} & \multirow{2}{*}{Encoder variants}  & \multirow{2}{*}{Decoder FLOPS} & \multicolumn{4}{c|}{AMOS} & \multicolumn{4}{c|}{FLARE'22} & \multicolumn{4}{c|}{BTCV} & \multirow{2}{*}{Avg} \\
\cline{4-15}
 &  &  & 1\% & 10\% & 100\% & Avg & 1\% & 10\% & 100\% & Avg &  1\% & 10\% & 100\% & Avg & \\
\hline
\multicolumn{2}{l|}{\textit{{\color{Gray} Decoder-driven reconstruction}}}  & & & & & & & & & & & & & \\
MAE & ViT-B/12$^{(1536)}$ & 16.44\ G & 58.84 & 75.74 & 84.80 & 73.56 & \textbf{63.33}  & 78.67 & 83.80 & 72.08 & 66.85 & 78.36 & 83.20 & 76.13 & 73.92 \\
\multicolumn{2}{l|}{\textit{{\color{Gray} Encoder-driven reconstruction}}}  & & & & & & & & & & & & & \\
Hi-End-MAE & ViT-B/12$^{(1536)}$ & 10.69\ G  & \textbf{60.03} & \textbf{75.83} & \textbf{84.97} & \textbf{73.61} & 63.22 & \textbf{80.58} & \textbf{84.19} & \textbf{74.00} & \textbf{69.59} & \textbf{78.56} & \textbf{84.53} & \textbf{77.56} & \textbf{75.05} \\

\Xhline{1px} 

\end{tabular}
}

\vspace{-2mm}
\label{tab:ablation1}
\end{table*}

\renewcommand{\multirowsetup}{\centering}  
\begin{table*}[t]
\caption{Evaluation of dense-decoding components with different variants encoder. We report the one-shot DSC score (\%) on six datasets. \textbf{val} (bold) / \underline{val} (underline) : top method / second method.}
\vspace{-2mm}
\centering
\resizebox{0.98\linewidth}{!}
{
\begin{tabular}{l  | cccc | c | cccccc | c}
\Xhline{1px} 
\multirow{2}{*}{Encoder variants} & \multicolumn{5}{c|}{Dense Decoding}  & \multicolumn{6}{c|}{Dataset} & \multirow{2}{*}{Avg} \\
\cline{2-6} \cline{7-12}
 & \multirow{1}{*}{w/o} &  \multirow{1}{*}{stage3} & \multirow{1}{*}{stage2} & \multirow{1}{*}{stage1} & FLOPS & BTCV & CT-ORG & Sliver07 & WORD & AMOS & FLARE'22 &  \\
\hline
 \multirow{3}{*}{ViT-B/16$^{(768)}$} & - & - & - &  - & - & 24.27 & 49.08 & 80.87 & 30.89 & 10.06 & 26.30 & 36.91 \\
 & \checkmark &  &  &  & 3.51\ G &  62.04 & 69.75 & 78.02 & 69.06 & 38.05 & 62.35 & 63.21 \\
 &  & \checkmark & \checkmark & \checkmark & 2.38\ G & 63.21 & 70.84 & 80.30 & 66.32 & 38.55 & 60.83 & 63.34 \\
\hline
 \multirow{3}{*}{ViT-B/16$^{(1536)}$} & - & - & - &  - & - & 27.52 & 46.43 & 76.16 & 31.07 & 7.47 & 23.41 & 35.34 \\
 & \checkmark &  &  &  & 6.41\ G & 61.94 & 69.37 & 83.92 & 67.23 & 38.52 & 59.30 & 63.38 \\
 &  & \checkmark & \checkmark & \checkmark & 4.39\ G & 62.15 & 70.66 & 89.71 & 69.05 & 37.70 & 63.27 & 65.42 \\
\hline
 \multirow{5}{*}{ViT-B/12$^{(1536)}$} & - & - & - &  - & - & 28.05 & 52.11 & 80.62 & 34.79 & 11.04 & 27.94 & 39.09 \\
 & \checkmark &  &  &  & 16.44\ G & 66.78 & \textbf{71.38} & \underline{88.76} & 71.68 & 39.62 & 63.33 & 66.92   \\
 &  &  &  & \checkmark & 14.52\ G & 64.98 & 69.12 & 88.29 & 71.33 & 40.78 & 62.28 & 66.13  \\
 &  &  & \checkmark & \checkmark & 12.61\ G & \underline{69.07} & \underline{71.28} & 86.46 & \underline{72.72} & \underline{42.39} & \textbf{64.41} & \underline{67.72}  \\
 &  & \checkmark & \checkmark & \checkmark & 10.69\ G & \textbf{69.59} & 71.09 & \textbf{91.88} & \textbf{73.52} & \textbf{46.21} & \underline{63.22} & \textbf{69.25} \\

\Xhline{1px} 

\end{tabular}
}

 \vspace{-2mm}
\label{tab:ablation2}
\end{table*}

\subsection{Analysis}
\noindent\textbf{Representation analysis.} Following previous work~\citep{effectiverank, theoretical}, we analyze representational capability by calculating the distributions of effective ranks and singular values~\citep{effectiverank} for the attention values ($V$) across each transformer layer in both the MAE~\citep{mae, mae3d} and Hi-End-MAE encoders. Except for input-related shallow layers, Hi-End-MAE demonstrates a broader range of effective ranks and higher singular values are positively correlated with stronger representational capacity. The results for all transformer layers are shown in Fig.\ref{fig:total_explain_er} and Fig.\ref{fig:total_explain_sv}. In terms of effective rank  (Fig.\ref{fig:total_explain_er}), Hi-End-MAE exhibits a notable distribution shift compared to MAE, exhibiting a broader range of feature representations. Furthermore, Hi-End-MAE shows a shift towards higher singular values (Fig.\ref{fig:total_explain_sv}), indicating a greater presence of high-variance components, which contribute to more detailed and diverse representations.


\noindent\textbf{Local attention patterns.} To intuitively demonstrate the specific representations learned by Hi-End-MAE, we visualize the slice-level (row 1) and volume-level (row 2) attention maps in Fig.\ref{fig:vis_attention}. These maps are generated by calculating the attention scores between a given query (highlighted in red boxes) and other patches. In contrast to decoder-driven methods, such as MAE, our encoder-driven Hi-End-MAE's self-attention maps focus more precisely on local patterns within specific organs, \textit{e.g.}, tubular attention in structures like the aorta, inferior vena cava, and veins, as well as clustered attention in the liver, kidneys, and spleen. Then, we visualize the different attention heads from 3th, 6th, 9th, and 12th ViT layers. As evidenced in Fig.\ref{fig:attn_map}, our Hi-End-MAE demonstrates markedly enhanced capability in anatomical attention across multiple ViT layers, producing more discriminative fine-grained feature essential for medical image segmentation. Empirical evidence indicates that Hi-End-MAE empowers individual ViT tokens to achieve rich aggregation of local medical features, transfer into strong performance gains for pixel-level medical downstream tasks.

\subsection{Ablations}

We first analyze the effects of decoder-driven and encoder-driven reconstruction on AMOS~\citep{amos}, FLARE'22~\citep{flare}, and BTCV~\citep{flare} datasets with 1\%, 10\%, and 100\% data proportions. Then, we conduct comprehensive ablation studies, across six datasets for one-shot segmentation, to evaluate dense decoder components, the mask ratios, and the scaling capabilities of ViT by Hi-End-MAE pre-training.

\noindent\textbf{Encoder-driven against Decoder-driven reconstruction.} We first explore the gains brought by our proposed encoder-driven reconstruction, as shown in Table~\ref{tab:ablation1}. Our proposed method achieves better results, \textit{e.g.}, 1.13\% average DSC improvement, and lower computational cost, \textit{e.g.}, 34.97\% computational cost reduced, than decoder-driven reconstruction across three datasets. This demonstrates that the encoder-driven reconstruction could enhance representation quality (Table~\ref{tab:ablation1}) and learn localized anatomical representations (Fig.~\ref{fig:total_explain_er}), benefiting various downstream tasks and highlighting the promising potential.

\noindent\textbf{Dense decoder components.} We further evaluate the effect of dense decoder number settings ($B$), \textit{e.g.}, by replacing self-attention mechanism in original decoder with cross-attention in Hi-End-MAE across various encoder variants. As shown in Table~\ref{tab:ablation2}, the results demonstrate that introducing the dense decoder in pre-training improves performance while reducing computational cost across all variants. Additionally, to assess the contribution of each block in different stages, we incrementally add dense decoder blocks from bottom to top in ViT-B/12$^{(1536)}$, achieving progressive gains in both performance and computational efficiency. With $B=3$, we achieve the highest performance (\textit{e.g.}, 69.25\% DSC) and the lowest computational cost (\textit{e.g.}, 10.69 GFLOPS). This suggests that progressive gains arise from the incremental addition of dense decoder blocks, which gradually reduces the decoder's workload and compels the encoder to learn higher-quality, localized anatomical representations.

\noindent\textbf{Scaling-up ViT in medical images.} We evaluate the scalability of Hi-End-MAE with respect to both data size and model size. As shown in Fig.{fig:vs}, with different data scales (\textit{e.g.}, 1K, 3K, 5K, 10K),  Hi-End-MAE learns better representations than other medical SSL methods, achieving superior and scalable downstream performance. Specifically, performance increases from 71.75\% to 78.21\% for 10\% fine-tuning and from 63.32\% to 69.25\% for one-shot segmentation as the data scale grows from 1K to 10K. Regarding model size, we compare ViT-B/16$^{(768)}$, ViT-B/16$^{(1536)}$ and ViT-B/12$^{(1536)}$ with same-settings for pre-training. The results, presented in Table~\ref{tab:ablation2}, show that Hi-End-MAE outperforms training from scratch (Table~\ref{tab:ablation2} row 1) and pre-training with MAE (Table~\ref{tab:ablation2} row 2), demonstrating better scalability, superior performance, and lower computational cost.

\noindent\textbf{Mask ratio.} We also conduct experiments with different mask ratios including 75\%, 80\%, 85\%, 90\% and 95\%. Results
are listed in Table~\ref{tab:maskratio}. Our findings show that a 75\% mask ratio yields better performance than higher mask ratios.

\section{Conclusion}
The necessity of low-level and high-quality localized anatomical representation learning prompts us to explore a novel architecture paradigm for fine-grained medical image pre-training. In this paper, we present Hi-End-MAE, a simple yet effective medical SSL framework. Different from previous decoder-driven reconstruction, Hi-End-MAE utilizes encoder-driven dense-decoding to gain high-quality medical representation. Specifically, it contains two parts: (1)  encoder-driven reconstruction utilizing decoder tokens to query visible encoded representations and (2) Hierarchical dense decoding performing densely bottom-up hierarchical decoding to learn informative anatomical representations. Extensive experiments demonstrate Hi-End-MAE brings significant performance leaps in downstream tasks and it reveals the encoder-driven reconstruction could learn strong localized anatomical representation across different ViT layers. We hope our encoder-driven paradigm could inspire more work to maximize the potential of masked image modeling in medical self-supervised learning tasks.

\section{Acknowledgments}
Supported by Natural Science Foundation of China under Grant 62271465, Suzhou Basic Research Program under Grant SYG202338, and Open Fund Project of Guangdong Academy of Medical Sciences, China (No. YKY-KF202206).

\bibliographystyle{model2-names.bst}\biboptions{authoryear}
\bibliography{main}

\begin{thebibliography}{65}
\expandafter\ifx\csname natexlab\endcsname\relax\def\natexlab#1{#1}\fi
\providecommand{\url}[1]{\texttt{#1}}
\providecommand{\href}[2]{#2}
\providecommand{\path}[1]{#1}
\providecommand{\DOIprefix}{doi:}
\providecommand{\ArXivprefix}{arXiv:}
\providecommand{\URLprefix}{URL: }
\providecommand{\Pubmedprefix}{pmid:}
\providecommand{\doi}[1]{\href{http://dx.doi.org/#1}{\path{#1}}}
\providecommand{\Pubmed}[1]{\href{pmid:#1}{\path{#1}}}
\providecommand{\bibinfo}[2]{#2}
\ifx\xfnm\relax \def\xfnm[#1]{\unskip,\space#1}\fi
\bibitem[{Assran et~al.(2022a)Assran, Balestriero, Duval, Bordes, Misra, Bojanowski, Vincent, Rabbat and Ballas}]{bias}
\bibinfo{author}{Assran, M.}, \bibinfo{author}{Balestriero, R.}, \bibinfo{author}{Duval, Q.}, \bibinfo{author}{Bordes, F.}, \bibinfo{author}{Misra, I.}, \bibinfo{author}{Bojanowski, P.}, \bibinfo{author}{Vincent, P.}, \bibinfo{author}{Rabbat, M.}, \bibinfo{author}{Ballas, N.}, \bibinfo{year}{2022}a.
\newblock \bibinfo{title}{The hidden uniform cluster prior in self-supervised learning}.
\newblock \bibinfo{journal}{arXiv preprint arXiv:2210.07277} .
\bibitem[{Assran et~al.(2022b)Assran, Caron, Misra, Bojanowski, Bordes, Vincent, Joulin, Rabbat and Ballas}]{highlevel}
\bibinfo{author}{Assran, M.}, \bibinfo{author}{Caron, M.}, \bibinfo{author}{Misra, I.}, \bibinfo{author}{Bojanowski, P.}, \bibinfo{author}{Bordes, F.}, \bibinfo{author}{Vincent, P.}, \bibinfo{author}{Joulin, A.}, \bibinfo{author}{Rabbat, M.}, \bibinfo{author}{Ballas, N.}, \bibinfo{year}{2022}b.
\newblock \bibinfo{title}{Masked siamese networks for label-efficient learning}, in: \bibinfo{booktitle}{European Conference on Computer Vision}, \bibinfo{organization}{Springer}. pp. \bibinfo{pages}{456--473}.
\bibitem[{Baid et~al.(2021)Baid, Ghodasara, Mohan, Bilello, Calabrese, Colak, Farahani, Kalpathy-Cramer, Kitamura, Pati et~al.}]{brats21}
\bibinfo{author}{Baid, U.}, \bibinfo{author}{Ghodasara, S.}, \bibinfo{author}{Mohan, S.}, \bibinfo{author}{Bilello, M.}, \bibinfo{author}{Calabrese, E.}, \bibinfo{author}{Colak, E.}, \bibinfo{author}{Farahani, K.}, \bibinfo{author}{Kalpathy-Cramer, J.}, \bibinfo{author}{Kitamura, F.C.}, \bibinfo{author}{Pati, S.}, et~al., \bibinfo{year}{2021}.
\newblock \bibinfo{title}{The rsna-asnr-miccai brats 2021 benchmark on brain tumor segmentation and radiogenomic classification}.
\newblock \bibinfo{journal}{arXiv preprint arXiv:2107.02314} .
\bibitem[{Bao et~al.(2021)Bao, Dong, Piao and Wei}]{beit}
\bibinfo{author}{Bao, H.}, \bibinfo{author}{Dong, L.}, \bibinfo{author}{Piao, S.}, \bibinfo{author}{Wei, F.}, \bibinfo{year}{2021}.
\newblock \bibinfo{title}{Beit: Bert pre-training of image transformers}.
\newblock \bibinfo{journal}{arXiv preprint arXiv:2106.08254} .
\bibitem[{Bilic et~al.(2023)Bilic, Christ, Li, Vorontsov, Ben-Cohen, Kaissis, Szeskin, Jacobs, Mamani, Chartrand et~al.}]{lits}
\bibinfo{author}{Bilic, P.}, \bibinfo{author}{Christ, P.}, \bibinfo{author}{Li, H.B.}, \bibinfo{author}{Vorontsov, E.}, \bibinfo{author}{Ben-Cohen, A.}, \bibinfo{author}{Kaissis, G.}, \bibinfo{author}{Szeskin, A.}, \bibinfo{author}{Jacobs, C.}, \bibinfo{author}{Mamani, G.E.H.}, \bibinfo{author}{Chartrand, G.}, et~al., \bibinfo{year}{2023}.
\newblock \bibinfo{title}{The liver tumor segmentation benchmark (lits)}.
\newblock \bibinfo{journal}{Medical Image Analysis} \bibinfo{volume}{84}, \bibinfo{pages}{102680}.
\bibitem[{Caron et~al.(2021)Caron, Touvron, Misra, J{\'e}gou, Mairal, Bojanowski and Joulin}]{dino}
\bibinfo{author}{Caron, M.}, \bibinfo{author}{Touvron, H.}, \bibinfo{author}{Misra, I.}, \bibinfo{author}{J{\'e}gou, H.}, \bibinfo{author}{Mairal, J.}, \bibinfo{author}{Bojanowski, P.}, \bibinfo{author}{Joulin, A.}, \bibinfo{year}{2021}.
\newblock \bibinfo{title}{Emerging properties in self-supervised vision transformers}, in: \bibinfo{booktitle}{Proceedings of the IEEE/CVF international conference on computer vision}, pp. \bibinfo{pages}{9650--9660}.
\bibitem[{Chen et~al.(2020)Chen, Kornblith, Norouzi and Hinton}]{simclr}
\bibinfo{author}{Chen, T.}, \bibinfo{author}{Kornblith, S.}, \bibinfo{author}{Norouzi, M.}, \bibinfo{author}{Hinton, G.}, \bibinfo{year}{2020}.
\newblock \bibinfo{title}{A simple framework for contrastive learning of visual representations}, in: \bibinfo{booktitle}{International conference on machine learning}, \bibinfo{organization}{PMLR}. pp. \bibinfo{pages}{1597--1607}.
\bibitem[{Chen et~al.(2024a)Chen, Ding, Wang, Xin, Mo, Wang, Han, Luo, Zeng and Wang}]{cae}
\bibinfo{author}{Chen, X.}, \bibinfo{author}{Ding, M.}, \bibinfo{author}{Wang, X.}, \bibinfo{author}{Xin, Y.}, \bibinfo{author}{Mo, S.}, \bibinfo{author}{Wang, Y.}, \bibinfo{author}{Han, S.}, \bibinfo{author}{Luo, P.}, \bibinfo{author}{Zeng, G.}, \bibinfo{author}{Wang, J.}, \bibinfo{year}{2024}a.
\newblock \bibinfo{title}{Context autoencoder for self-supervised representation learning}.
\newblock \bibinfo{journal}{International Journal of Computer Vision} \bibinfo{volume}{132}, \bibinfo{pages}{208--223}.
\bibitem[{Chen et~al.(2024b)Chen, Ding, Wang, Xin, Mo, Wang, Han, Luo, Zeng and Wang}]{cmae}
\bibinfo{author}{Chen, X.}, \bibinfo{author}{Ding, M.}, \bibinfo{author}{Wang, X.}, \bibinfo{author}{Xin, Y.}, \bibinfo{author}{Mo, S.}, \bibinfo{author}{Wang, Y.}, \bibinfo{author}{Han, S.}, \bibinfo{author}{Luo, P.}, \bibinfo{author}{Zeng, G.}, \bibinfo{author}{Wang, J.}, \bibinfo{year}{2024}b.
\newblock \bibinfo{title}{Context autoencoder for self-supervised representation learning}.
\newblock \bibinfo{journal}{International Journal of Computer Vision} \bibinfo{volume}{132}, \bibinfo{pages}{208--223}.
\bibitem[{Chen et~al.(2023a)Chen, Agarwal, Aggarwal, Safta, Balan and Brown}]{mim}
\bibinfo{author}{Chen, Z.}, \bibinfo{author}{Agarwal, D.}, \bibinfo{author}{Aggarwal, K.}, \bibinfo{author}{Safta, W.}, \bibinfo{author}{Balan, M.M.}, \bibinfo{author}{Brown, K.}, \bibinfo{year}{2023}a.
\newblock \bibinfo{title}{Masked image modeling advances 3d medical image analysis}, in: \bibinfo{booktitle}{Proceedings of the IEEE/CVF Winter Conference on Applications of Computer Vision}, pp. \bibinfo{pages}{1970--1980}.
\bibitem[{Chen et~al.(2023b)Chen, Agarwal, Aggarwal, Safta, Balan and Brown}]{benchmark}
\bibinfo{author}{Chen, Z.}, \bibinfo{author}{Agarwal, D.}, \bibinfo{author}{Aggarwal, K.}, \bibinfo{author}{Safta, W.}, \bibinfo{author}{Balan, M.M.}, \bibinfo{author}{Brown, K.}, \bibinfo{year}{2023}b.
\newblock \bibinfo{title}{Masked image modeling advances 3d medical image analysis}, in: \bibinfo{booktitle}{Proceedings of the IEEE/CVF Winter Conference on Applications of Computer Vision}, pp. \bibinfo{pages}{1970--1980}.
\bibitem[{Clark et~al.(2013)Clark, Vendt, Smith, Freymann, Kirby, Koppel, Moore, Phillips, Maffitt, Pringle et~al.}]{flare_center}
\bibinfo{author}{Clark, K.}, \bibinfo{author}{Vendt, B.}, \bibinfo{author}{Smith, K.}, \bibinfo{author}{Freymann, J.}, \bibinfo{author}{Kirby, J.}, \bibinfo{author}{Koppel, P.}, \bibinfo{author}{Moore, S.}, \bibinfo{author}{Phillips, S.}, \bibinfo{author}{Maffitt, D.}, \bibinfo{author}{Pringle, M.}, et~al., \bibinfo{year}{2013}.
\newblock \bibinfo{title}{The cancer imaging archive (tcia): maintaining and operating a public information repository}.
\newblock \bibinfo{journal}{Journal of digital imaging} \bibinfo{volume}{26}, \bibinfo{pages}{1045--1057}.
\bibitem[{Dong et~al.(2023)Dong, Bao, Zhang, Chen, Zhang, Yuan, Chen, Wen, Yu and Guo}]{peco}
\bibinfo{author}{Dong, X.}, \bibinfo{author}{Bao, J.}, \bibinfo{author}{Zhang, T.}, \bibinfo{author}{Chen, D.}, \bibinfo{author}{Zhang, W.}, \bibinfo{author}{Yuan, L.}, \bibinfo{author}{Chen, D.}, \bibinfo{author}{Wen, F.}, \bibinfo{author}{Yu, N.}, \bibinfo{author}{Guo, B.}, \bibinfo{year}{2023}.
\newblock \bibinfo{title}{Peco: Perceptual codebook for bert pre-training of vision transformers}, in: \bibinfo{booktitle}{Proceedings of the AAAI Conference on Artificial Intelligence}, pp. \bibinfo{pages}{552--560}.
\bibitem[{Dosovitskiy(2020)}]{vit}
\bibinfo{author}{Dosovitskiy, A.}, \bibinfo{year}{2020}.
\newblock \bibinfo{title}{An image is worth 16x16 words: Transformers for image recognition at scale}.
\newblock \bibinfo{journal}{arXiv preprint arXiv:2010.11929} .
\bibitem[{Goncharov et~al.(2023)Goncharov, Soboleva, Kurmukov, Pisov and Belyaev}]{vox2vec}
\bibinfo{author}{Goncharov, M.}, \bibinfo{author}{Soboleva, V.}, \bibinfo{author}{Kurmukov, A.}, \bibinfo{author}{Pisov, M.}, \bibinfo{author}{Belyaev, M.}, \bibinfo{year}{2023}.
\newblock \bibinfo{title}{vox2vec: A framework for self-supervised contrastive learning of voxel-level representations in medical images}, in: \bibinfo{booktitle}{International Conference on Medical Image Computing and Computer-Assisted Intervention}, \bibinfo{organization}{Springer}. pp. \bibinfo{pages}{605--614}.
\bibitem[{Gowda and Clifton(2024)}]{ratio3}
\bibinfo{author}{Gowda, S.N.}, \bibinfo{author}{Clifton, D.A.}, \bibinfo{year}{2024}.
\newblock \bibinfo{title}{Masks and manuscripts: Advancing medical pre-training with end-to-end masking and narrative structuring}, in: \bibinfo{booktitle}{International Conference on Medical Image Computing and Computer-Assisted Intervention}, \bibinfo{organization}{Springer}. pp. \bibinfo{pages}{426--436}.
\bibitem[{Haghighi et~al.(2021)Haghighi, Taher, Zhou, Gotway and Liang}]{transvw}
\bibinfo{author}{Haghighi, F.}, \bibinfo{author}{Taher, M.R.H.}, \bibinfo{author}{Zhou, Z.}, \bibinfo{author}{Gotway, M.B.}, \bibinfo{author}{Liang, J.}, \bibinfo{year}{2021}.
\newblock \bibinfo{title}{Transferable visual words: Exploiting the semantics of anatomical patterns for self-supervised learning}.
\newblock \bibinfo{journal}{IEEE transactions on medical imaging} \bibinfo{volume}{40}, \bibinfo{pages}{2857--2868}.
\bibitem[{Hatamizadeh et~al.(2022)Hatamizadeh, Tang, Nath, Yang, Myronenko, Landman, Roth and Xu}]{unetr}
\bibinfo{author}{Hatamizadeh, A.}, \bibinfo{author}{Tang, Y.}, \bibinfo{author}{Nath, V.}, \bibinfo{author}{Yang, D.}, \bibinfo{author}{Myronenko, A.}, \bibinfo{author}{Landman, B.}, \bibinfo{author}{Roth, H.R.}, \bibinfo{author}{Xu, D.}, \bibinfo{year}{2022}.
\newblock \bibinfo{title}{Unetr: Transformers for 3d medical image segmentation}, in: \bibinfo{booktitle}{Proceedings of the IEEE/CVF winter conference on applications of computer vision}, pp. \bibinfo{pages}{574--584}.
\bibitem[{He et~al.(2022)He, Chen, Xie, Li, Doll{\'a}r and Girshick}]{mae}
\bibinfo{author}{He, K.}, \bibinfo{author}{Chen, X.}, \bibinfo{author}{Xie, S.}, \bibinfo{author}{Li, Y.}, \bibinfo{author}{Doll{\'a}r, P.}, \bibinfo{author}{Girshick, R.}, \bibinfo{year}{2022}.
\newblock \bibinfo{title}{Masked autoencoders are scalable vision learners}, in: \bibinfo{booktitle}{Proceedings of the IEEE/CVF conference on computer vision and pattern recognition}, pp. \bibinfo{pages}{16000--16009}.
\bibitem[{He et~al.(2020a)He, Fan, Wu, Xie and Girshick}]{moco}
\bibinfo{author}{He, K.}, \bibinfo{author}{Fan, H.}, \bibinfo{author}{Wu, Y.}, \bibinfo{author}{Xie, S.}, \bibinfo{author}{Girshick, R.}, \bibinfo{year}{2020}a.
\newblock \bibinfo{title}{Momentum contrast for unsupervised visual representation learning}, in: \bibinfo{booktitle}{Proceedings of the IEEE/CVF conference on computer vision and pattern recognition}, pp. \bibinfo{pages}{9729--9738}.
\bibitem[{He et~al.(2023)He, Yang, Ge, Chen, Coatrieux, Wang and Li}]{gvsl}
\bibinfo{author}{He, Y.}, \bibinfo{author}{Yang, G.}, \bibinfo{author}{Ge, R.}, \bibinfo{author}{Chen, Y.}, \bibinfo{author}{Coatrieux, J.L.}, \bibinfo{author}{Wang, B.}, \bibinfo{author}{Li, S.}, \bibinfo{year}{2023}.
\newblock \bibinfo{title}{Geometric visual similarity learning in 3d medical image self-supervised pre-training}, in: \bibinfo{booktitle}{Proceedings of the IEEE/CVF Conference on Computer Vision and Pattern Recognition}, pp. \bibinfo{pages}{9538--9547}.
\bibitem[{He et~al.(2020b)He, Yang, Yang, Chen, Kong, Wu, Tang, Zhu, Dillenseger, Shao et~al.}]{kipa2}
\bibinfo{author}{He, Y.}, \bibinfo{author}{Yang, G.}, \bibinfo{author}{Yang, J.}, \bibinfo{author}{Chen, Y.}, \bibinfo{author}{Kong, Y.}, \bibinfo{author}{Wu, J.}, \bibinfo{author}{Tang, L.}, \bibinfo{author}{Zhu, X.}, \bibinfo{author}{Dillenseger, J.L.}, \bibinfo{author}{Shao, P.}, et~al., \bibinfo{year}{2020}b.
\newblock \bibinfo{title}{Dense biased networks with deep priori anatomy and hard region adaptation: Semi-supervised learning for fine renal artery segmentation}.
\newblock \bibinfo{journal}{Medical image analysis} \bibinfo{volume}{63}, \bibinfo{pages}{101722}.
\bibitem[{He et~al.(2021)He, Yang, Yang, Ge, Kong, Zhu, Zhang, Shao, Shu, Dillenseger et~al.}]{kipa1}
\bibinfo{author}{He, Y.}, \bibinfo{author}{Yang, G.}, \bibinfo{author}{Yang, J.}, \bibinfo{author}{Ge, R.}, \bibinfo{author}{Kong, Y.}, \bibinfo{author}{Zhu, X.}, \bibinfo{author}{Zhang, S.}, \bibinfo{author}{Shao, P.}, \bibinfo{author}{Shu, H.}, \bibinfo{author}{Dillenseger, J.L.}, et~al., \bibinfo{year}{2021}.
\newblock \bibinfo{title}{Meta grayscale adaptive network for 3d integrated renal structures segmentation}.
\newblock \bibinfo{journal}{Medical image analysis} \bibinfo{volume}{71}, \bibinfo{pages}{102055}.
\bibitem[{Heimann et~al.(2009)Heimann, Van~Ginneken, Styner, Arzhaeva, Aurich, Bauer, Beck, Becker, Beichel, Bekes et~al.}]{sliver07}
\bibinfo{author}{Heimann, T.}, \bibinfo{author}{Van~Ginneken, B.}, \bibinfo{author}{Styner, M.A.}, \bibinfo{author}{Arzhaeva, Y.}, \bibinfo{author}{Aurich, V.}, \bibinfo{author}{Bauer, C.}, \bibinfo{author}{Beck, A.}, \bibinfo{author}{Becker, C.}, \bibinfo{author}{Beichel, R.}, \bibinfo{author}{Bekes, G.}, et~al., \bibinfo{year}{2009}.
\newblock \bibinfo{title}{Comparison and evaluation of methods for liver segmentation from ct datasets}.
\newblock \bibinfo{journal}{IEEE transactions on medical imaging} \bibinfo{volume}{28}, \bibinfo{pages}{1251--1265}.
\bibitem[{Huang et~al.(2024)Huang, Li, Zhou, Yang, Liu, Liang, Zheng, Zhang and Wang}]{ratio1}
\bibinfo{author}{Huang, W.}, \bibinfo{author}{Li, C.}, \bibinfo{author}{Zhou, H.Y.}, \bibinfo{author}{Yang, H.}, \bibinfo{author}{Liu, J.}, \bibinfo{author}{Liang, Y.}, \bibinfo{author}{Zheng, H.}, \bibinfo{author}{Zhang, S.}, \bibinfo{author}{Wang, S.}, \bibinfo{year}{2024}.
\newblock \bibinfo{title}{Enhancing representation in radiography-reports foundation model: A granular alignment algorithm using masked contrastive learning}.
\newblock \bibinfo{journal}{Nature Communications} \bibinfo{volume}{15}, \bibinfo{pages}{7620}.
\bibitem[{Isensee et~al.(2024)Isensee, Wald, Ulrich, Baumgartner, Roy, Maier-Hein and Jaeger}]{ratio4}
\bibinfo{author}{Isensee, F.}, \bibinfo{author}{Wald, T.}, \bibinfo{author}{Ulrich, C.}, \bibinfo{author}{Baumgartner, M.}, \bibinfo{author}{Roy, S.}, \bibinfo{author}{Maier-Hein, K.}, \bibinfo{author}{Jaeger, P.F.}, \bibinfo{year}{2024}.
\newblock \bibinfo{title}{nnu-net revisited: A call for rigorous validation in 3d medical image segmentation}, in: \bibinfo{booktitle}{International Conference on Medical Image Computing and Computer-Assisted Intervention}, \bibinfo{organization}{Springer}. pp. \bibinfo{pages}{488--498}.
\bibitem[{Ji et~al.(2022)Ji, Bai, Ge, Yang, Zhu, Zhang, Li, Zhanng, Ma, Wan et~al.}]{amos}
\bibinfo{author}{Ji, Y.}, \bibinfo{author}{Bai, H.}, \bibinfo{author}{Ge, C.}, \bibinfo{author}{Yang, J.}, \bibinfo{author}{Zhu, Y.}, \bibinfo{author}{Zhang, R.}, \bibinfo{author}{Li, Z.}, \bibinfo{author}{Zhanng, L.}, \bibinfo{author}{Ma, W.}, \bibinfo{author}{Wan, X.}, et~al., \bibinfo{year}{2022}.
\newblock \bibinfo{title}{Amos: A large-scale abdominal multi-organ benchmark for versatile medical image segmentation}.
\newblock \bibinfo{journal}{Advances in neural information processing systems} \bibinfo{volume}{35}, \bibinfo{pages}{36722--36732}.
\bibitem[{Kavur et~al.(2021)Kavur, Gezer, Bar{\i}{\c{s}}, Aslan, Conze, Groza, Pham, Chatterjee, Ernst, {\"O}zkan et~al.}]{chaos}
\bibinfo{author}{Kavur, A.E.}, \bibinfo{author}{Gezer, N.S.}, \bibinfo{author}{Bar{\i}{\c{s}}, M.}, \bibinfo{author}{Aslan, S.}, \bibinfo{author}{Conze, P.H.}, \bibinfo{author}{Groza, V.}, \bibinfo{author}{Pham, D.D.}, \bibinfo{author}{Chatterjee, S.}, \bibinfo{author}{Ernst, P.}, \bibinfo{author}{{\"O}zkan, S.}, et~al., \bibinfo{year}{2021}.
\newblock \bibinfo{title}{Chaos challenge-combined (ct-mr) healthy abdominal organ segmentation}.
\newblock \bibinfo{journal}{Medical Image Analysis} \bibinfo{volume}{69}, \bibinfo{pages}{101950}.
\bibitem[{Landman et~al.(2015)Landman, Xu, Igelsias, Styner, Langerak and Klein}]{btcv}
\bibinfo{author}{Landman, B.}, \bibinfo{author}{Xu, Z.}, \bibinfo{author}{Igelsias, J.}, \bibinfo{author}{Styner, M.}, \bibinfo{author}{Langerak, T.}, \bibinfo{author}{Klein, A.}, \bibinfo{year}{2015}.
\newblock \bibinfo{title}{Miccai multi-atlas labeling beyond the cranial vault--workshop and challenge}, in: \bibinfo{booktitle}{Proc. MICCAI Multi-Atlas Labeling Beyond Cranial Vault—Workshop Challenge}, p.~\bibinfo{pages}{12}.
\bibitem[{Lay et~al.(2013)Lay, Birkbeck, Zhang and Kevin~Zhou}]{lay2013rapid}
\bibinfo{author}{Lay, N.}, \bibinfo{author}{Birkbeck, N.}, \bibinfo{author}{Zhang, J.}, \bibinfo{author}{Kevin~Zhou, S.}, \bibinfo{year}{2013}.
\newblock \bibinfo{title}{Rapid multi-organ segmentation using context integration and discriminative models}, in: \bibinfo{booktitle}{International Conference on Information Processing in Medical Imaging}, \bibinfo{organization}{Springer, Berlin, Heidelberg}. pp. \bibinfo{pages}{450--462}.
\bibitem[{Li et~al.(2023)Li, Zhang, Wang, Wu, Wu, Liu, Xia, Tan, Liu, Sun et~al.}]{survey}
\bibinfo{author}{Li, S.}, \bibinfo{author}{Zhang, L.}, \bibinfo{author}{Wang, Z.}, \bibinfo{author}{Wu, D.}, \bibinfo{author}{Wu, L.}, \bibinfo{author}{Liu, Z.}, \bibinfo{author}{Xia, J.}, \bibinfo{author}{Tan, C.}, \bibinfo{author}{Liu, Y.}, \bibinfo{author}{Sun, B.}, et~al., \bibinfo{year}{2023}.
\newblock \bibinfo{title}{Masked modeling for self-supervised representation learning on vision and beyond}.
\newblock \bibinfo{journal}{arXiv preprint arXiv:2401.00897} .
\bibitem[{Li et~al.(2024)Li, Qu, Chen, Bassi, Shi, Lai, Yu, Xue, Chen, Lin et~al.}]{atlas10}
\bibinfo{author}{Li, W.}, \bibinfo{author}{Qu, C.}, \bibinfo{author}{Chen, X.}, \bibinfo{author}{Bassi, P.R.}, \bibinfo{author}{Shi, Y.}, \bibinfo{author}{Lai, Y.}, \bibinfo{author}{Yu, Q.}, \bibinfo{author}{Xue, H.}, \bibinfo{author}{Chen, Y.}, \bibinfo{author}{Lin, X.}, et~al., \bibinfo{year}{2024}.
\newblock \bibinfo{title}{Abdomenatlas: A large-scale, detailed-annotated, \& multi-center dataset for efficient transfer learning and open algorithmic benchmarking}.
\newblock \bibinfo{journal}{Medical Image Analysis} \bibinfo{volume}{97}, \bibinfo{pages}{103285}.
\bibitem[{Luo et~al.(2022)Luo, Liao, Xiao, Chen, Song, Zhang, Li, Metaxas, Wang and Zhang}]{word}
\bibinfo{author}{Luo, X.}, \bibinfo{author}{Liao, W.}, \bibinfo{author}{Xiao, J.}, \bibinfo{author}{Chen, J.}, \bibinfo{author}{Song, T.}, \bibinfo{author}{Zhang, X.}, \bibinfo{author}{Li, K.}, \bibinfo{author}{Metaxas, D.N.}, \bibinfo{author}{Wang, G.}, \bibinfo{author}{Zhang, S.}, \bibinfo{year}{2022}.
\newblock \bibinfo{title}{Word: A large scale dataset, benchmark and clinical applicable study for abdominal organ segmentation from ct image}.
\newblock \bibinfo{journal}{Medical Image Analysis} \bibinfo{volume}{82}, \bibinfo{pages}{102642}.
\bibitem[{Ma et~al.(2021a)Ma, Wang, An, Ge, Yu, Chen, Zhu, Dong, He, He et~al.}]{convid}
\bibinfo{author}{Ma, J.}, \bibinfo{author}{Wang, Y.}, \bibinfo{author}{An, X.}, \bibinfo{author}{Ge, C.}, \bibinfo{author}{Yu, Z.}, \bibinfo{author}{Chen, J.}, \bibinfo{author}{Zhu, Q.}, \bibinfo{author}{Dong, G.}, \bibinfo{author}{He, J.}, \bibinfo{author}{He, Z.}, et~al., \bibinfo{year}{2021}a.
\newblock \bibinfo{title}{Toward data-efficient learning: A benchmark for covid-19 ct lung and infection segmentation}.
\newblock \bibinfo{journal}{Medical physics} \bibinfo{volume}{48}, \bibinfo{pages}{1197--1210}.
\bibitem[{Ma et~al.(2023)Ma, Zhang, Gu, Ge, Ma, Young, Zhu, Meng, Yang, Huang et~al.}]{flare}
\bibinfo{author}{Ma, J.}, \bibinfo{author}{Zhang, Y.}, \bibinfo{author}{Gu, S.}, \bibinfo{author}{Ge, C.}, \bibinfo{author}{Ma, S.}, \bibinfo{author}{Young, A.}, \bibinfo{author}{Zhu, C.}, \bibinfo{author}{Meng, K.}, \bibinfo{author}{Yang, X.}, \bibinfo{author}{Huang, Z.}, et~al., \bibinfo{year}{2023}.
\newblock \bibinfo{title}{Unleashing the strengths of unlabeled data in pan-cancer abdominal organ quantification: the flare22 challenge}.
\newblock \bibinfo{journal}{arXiv preprint arXiv:2308.05862} .
\bibitem[{Ma et~al.(2021b)Ma, Zhang, Gu, Zhu, Ge, Zhang, An, Wang, Wang, Liu et~al.}]{abd1k}
\bibinfo{author}{Ma, J.}, \bibinfo{author}{Zhang, Y.}, \bibinfo{author}{Gu, S.}, \bibinfo{author}{Zhu, C.}, \bibinfo{author}{Ge, C.}, \bibinfo{author}{Zhang, Y.}, \bibinfo{author}{An, X.}, \bibinfo{author}{Wang, C.}, \bibinfo{author}{Wang, Q.}, \bibinfo{author}{Liu, X.}, et~al., \bibinfo{year}{2021}b.
\newblock \bibinfo{title}{Abdomenct-1k: Is abdominal organ segmentation a solved problem?}
\newblock \bibinfo{journal}{IEEE Transactions on Pattern Analysis and Machine Intelligence} \bibinfo{volume}{44}, \bibinfo{pages}{6695--6714}.
\bibitem[{Park et~al.(2024)Park, Jung, Lee, Ye and Lee}]{er2}
\bibinfo{author}{Park, G.Y.}, \bibinfo{author}{Jung, C.}, \bibinfo{author}{Lee, S.}, \bibinfo{author}{Ye, J.C.}, \bibinfo{author}{Lee, S.W.}, \bibinfo{year}{2024}.
\newblock \bibinfo{title}{Self-supervised debiasing using low rank regularization}, in: \bibinfo{booktitle}{Proceedings of the IEEE/CVF Conference on Computer Vision and Pattern Recognition}, pp. \bibinfo{pages}{12395--12405}.
\bibitem[{Pathak et~al.(2016)Pathak, Krahenbuhl, Donahue, Darrell and Efros}]{inpaint}
\bibinfo{author}{Pathak, D.}, \bibinfo{author}{Krahenbuhl, P.}, \bibinfo{author}{Donahue, J.}, \bibinfo{author}{Darrell, T.}, \bibinfo{author}{Efros, A.A.}, \bibinfo{year}{2016}.
\newblock \bibinfo{title}{Context encoders: Feature learning by inpainting}, in: \bibinfo{booktitle}{Proceedings of the IEEE conference on computer vision and pattern recognition}, pp. \bibinfo{pages}{2536--2544}.
\bibitem[{Podobnik et~al.(2023)Podobnik, Strojan, Peterlin, Ibragimov and Vrtovec}]{han}
\bibinfo{author}{Podobnik, G.}, \bibinfo{author}{Strojan, P.}, \bibinfo{author}{Peterlin, P.}, \bibinfo{author}{Ibragimov, B.}, \bibinfo{author}{Vrtovec, T.}, \bibinfo{year}{2023}.
\newblock \bibinfo{title}{Han-seg: The head and neck organ-at-risk ct and mr segmentation dataset}.
\newblock \bibinfo{journal}{Medical physics} \bibinfo{volume}{50}, \bibinfo{pages}{1917--1927}.
\bibitem[{Rister et~al.(2020)Rister, Yi, Shivakumar, Nobashi and Rubin}]{ctorg}
\bibinfo{author}{Rister, B.}, \bibinfo{author}{Yi, D.}, \bibinfo{author}{Shivakumar, K.}, \bibinfo{author}{Nobashi, T.}, \bibinfo{author}{Rubin, D.L.}, \bibinfo{year}{2020}.
\newblock \bibinfo{title}{Ct-org, a new dataset for multiple organ segmentation in computed tomography}.
\newblock \bibinfo{journal}{Scientific Data} \bibinfo{volume}{7}, \bibinfo{pages}{381}.
\bibitem[{Ronneberger et~al.(2015)Ronneberger, Fischer and Brox}]{unet}
\bibinfo{author}{Ronneberger, O.}, \bibinfo{author}{Fischer, P.}, \bibinfo{author}{Brox, T.}, \bibinfo{year}{2015}.
\newblock \bibinfo{title}{U-net: Convolutional networks for biomedical image segmentation}, in: \bibinfo{booktitle}{Medical image computing and computer-assisted intervention--MICCAI 2015: 18th international conference, Munich, Germany, October 5-9, 2015, proceedings, part III 18}, \bibinfo{organization}{Springer}. pp. \bibinfo{pages}{234--241}.
\bibitem[{Roth et~al.(2016)Roth, Farag, Turkbey, Lu, Liu and Summers}]{pan}
\bibinfo{author}{Roth, H.}, \bibinfo{author}{Farag, A.}, \bibinfo{author}{Turkbey, E.B.}, \bibinfo{author}{Lu, L.}, \bibinfo{author}{Liu, J.}, \bibinfo{author}{Summers, R.M.}, \bibinfo{year}{2016}.
\newblock \bibinfo{title}{Data from pancreas-ct.}
\newblock \bibinfo{journal}{The Cancer Imaging Archive.} .
\bibitem[{Roy and Vetterli(2007)}]{effectiverank}
\bibinfo{author}{Roy, O.}, \bibinfo{author}{Vetterli, M.}, \bibinfo{year}{2007}.
\newblock \bibinfo{title}{The effective rank: A measure of effective dimensionality}, in: \bibinfo{booktitle}{2007 15th European signal processing conference}, \bibinfo{organization}{IEEE}. pp. \bibinfo{pages}{606--610}.
\bibitem[{Roy et~al.(2023)Roy, Koehler, Ulrich, Baumgartner, Petersen, Isensee, Jaeger and Maier-Hein}]{mednext}
\bibinfo{author}{Roy, S.}, \bibinfo{author}{Koehler, G.}, \bibinfo{author}{Ulrich, C.}, \bibinfo{author}{Baumgartner, M.}, \bibinfo{author}{Petersen, J.}, \bibinfo{author}{Isensee, F.}, \bibinfo{author}{Jaeger, P.F.}, \bibinfo{author}{Maier-Hein, K.H.}, \bibinfo{year}{2023}.
\newblock \bibinfo{title}{Mednext: transformer-driven scaling of convnets for medical image segmentation}, in: \bibinfo{booktitle}{International Conference on Medical Image Computing and Computer-Assisted Intervention}, \bibinfo{organization}{Springer}. pp. \bibinfo{pages}{405--415}.
\bibitem[{Setio et~al.(2017)Setio, Traverso, De~Bel, Berens, Van Den~Bogaard, Cerello, Chen, Dou, Fantacci, Geurts et~al.}]{luna}
\bibinfo{author}{Setio, A.A.A.}, \bibinfo{author}{Traverso, A.}, \bibinfo{author}{De~Bel, T.}, \bibinfo{author}{Berens, M.S.}, \bibinfo{author}{Van Den~Bogaard, C.}, \bibinfo{author}{Cerello, P.}, \bibinfo{author}{Chen, H.}, \bibinfo{author}{Dou, Q.}, \bibinfo{author}{Fantacci, M.E.}, \bibinfo{author}{Geurts, B.}, et~al., \bibinfo{year}{2017}.
\newblock \bibinfo{title}{Validation, comparison, and combination of algorithms for automatic detection of pulmonary nodules in computed tomography images: the luna16 challenge}.
\newblock \bibinfo{journal}{Medical image analysis} \bibinfo{volume}{42}, \bibinfo{pages}{1--13}.
\bibitem[{Shao et~al.(2011)Shao, Qin, Yin, Meng, Ju, Li, Lv, Zhang and Xu}]{kipa3}
\bibinfo{author}{Shao, P.}, \bibinfo{author}{Qin, C.}, \bibinfo{author}{Yin, C.}, \bibinfo{author}{Meng, X.}, \bibinfo{author}{Ju, X.}, \bibinfo{author}{Li, J.}, \bibinfo{author}{Lv, Q.}, \bibinfo{author}{Zhang, W.}, \bibinfo{author}{Xu, Z.}, \bibinfo{year}{2011}.
\newblock \bibinfo{title}{Laparoscopic partial nephrectomy with segmental renal artery clamping: technique and clinical outcomes}.
\newblock \bibinfo{journal}{European urology} \bibinfo{volume}{59}, \bibinfo{pages}{849--855}.
\bibitem[{Shao et~al.(2012)Shao, Tang, Li, Xu, Qin, Cao, Ju, Meng, Lv, Li et~al.}]{kipa4}
\bibinfo{author}{Shao, P.}, \bibinfo{author}{Tang, L.}, \bibinfo{author}{Li, P.}, \bibinfo{author}{Xu, Y.}, \bibinfo{author}{Qin, C.}, \bibinfo{author}{Cao, Q.}, \bibinfo{author}{Ju, X.}, \bibinfo{author}{Meng, X.}, \bibinfo{author}{Lv, Q.}, \bibinfo{author}{Li, J.}, et~al., \bibinfo{year}{2012}.
\newblock \bibinfo{title}{Precise segmental renal artery clamping under the guidance of dual-source computed tomography angiography during laparoscopic partial nephrectomy}.
\newblock \bibinfo{journal}{European urology} \bibinfo{volume}{62}, \bibinfo{pages}{1001--1008}.
\bibitem[{Taleb et~al.(2020)Taleb, Loetzsch, Danz, Severin, Gaertner, Bergner and Lippert}]{ssl3d}
\bibinfo{author}{Taleb, A.}, \bibinfo{author}{Loetzsch, W.}, \bibinfo{author}{Danz, N.}, \bibinfo{author}{Severin, J.}, \bibinfo{author}{Gaertner, T.}, \bibinfo{author}{Bergner, B.}, \bibinfo{author}{Lippert, C.}, \bibinfo{year}{2020}.
\newblock \bibinfo{title}{3d self-supervised methods for medical imaging}.
\newblock \bibinfo{journal}{Advances in neural information processing systems} \bibinfo{volume}{33}, \bibinfo{pages}{18158--18172}.
\bibitem[{Tang et~al.(2024)Tang, Xu, Yao, Fu, Quan, Zhu, Liu and Zhou}]{hyspark}
\bibinfo{author}{Tang, F.}, \bibinfo{author}{Xu, R.}, \bibinfo{author}{Yao, Q.}, \bibinfo{author}{Fu, X.}, \bibinfo{author}{Quan, Q.}, \bibinfo{author}{Zhu, H.}, \bibinfo{author}{Liu, Z.}, \bibinfo{author}{Zhou, S.K.}, \bibinfo{year}{2024}.
\newblock \bibinfo{title}{Hyspark: Hybrid sparse masking for large scale medical image pre-training}, in: \bibinfo{booktitle}{International Conference on Medical Image Computing and Computer-Assisted Intervention}, \bibinfo{organization}{Springer}. pp. \bibinfo{pages}{330--340}.
\bibitem[{Tang et~al.(2022)Tang, Yang, Li, Roth, Landman, Xu, Nath and Hatamizadeh}]{swinunetr}
\bibinfo{author}{Tang, Y.}, \bibinfo{author}{Yang, D.}, \bibinfo{author}{Li, W.}, \bibinfo{author}{Roth, H.R.}, \bibinfo{author}{Landman, B.}, \bibinfo{author}{Xu, D.}, \bibinfo{author}{Nath, V.}, \bibinfo{author}{Hatamizadeh, A.}, \bibinfo{year}{2022}.
\newblock \bibinfo{title}{Self-supervised pre-training of swin transformers for 3d medical image analysis}, in: \bibinfo{booktitle}{Proceedings of the IEEE/CVF conference on computer vision and pattern recognition}, pp. \bibinfo{pages}{20730--20740}.
\bibitem[{Tao et~al.(2020)Tao, Li, Zhou, Ma and Zheng}]{rotate}
\bibinfo{author}{Tao, X.}, \bibinfo{author}{Li, Y.}, \bibinfo{author}{Zhou, W.}, \bibinfo{author}{Ma, K.}, \bibinfo{author}{Zheng, Y.}, \bibinfo{year}{2020}.
\newblock \bibinfo{title}{Revisiting rubik’s cube: Self-supervised learning with volume-wise transformation for 3d medical image segmentation}, in: \bibinfo{booktitle}{Medical Image Computing and Computer Assisted Intervention--MICCAI 2020: 23rd International Conference, Lima, Peru, October 4--8, 2020, Proceedings, Part IV 23}, \bibinfo{organization}{Springer}. pp. \bibinfo{pages}{238--248}.
\bibitem[{Tian et~al.(2023)Tian, Jiang, Diao, Lin, Wang and Yuan}]{spark}
\bibinfo{author}{Tian, K.}, \bibinfo{author}{Jiang, Y.}, \bibinfo{author}{Diao, Q.}, \bibinfo{author}{Lin, C.}, \bibinfo{author}{Wang, L.}, \bibinfo{author}{Yuan, Z.}, \bibinfo{year}{2023}.
\newblock \bibinfo{title}{Designing bert for convolutional networks: Sparse and hierarchical masked modeling}.
\newblock \bibinfo{journal}{arXiv preprint arXiv:2301.03580} .
\bibitem[{Wu et~al.(2024)Wu, Zhuang and Chen}]{voco}
\bibinfo{author}{Wu, L.}, \bibinfo{author}{Zhuang, J.}, \bibinfo{author}{Chen, H.}, \bibinfo{year}{2024}.
\newblock \bibinfo{title}{Voco: A simple-yet-effective volume contrastive learning framework for 3d medical image analysis}, in: \bibinfo{booktitle}{Proceedings of the IEEE/CVF Conference on Computer Vision and Pattern Recognition}, pp. \bibinfo{pages}{22873--22882}.
\bibitem[{Xie et~al.(2022a)Xie, Zhang, Xia and Wu}]{unimiss}
\bibinfo{author}{Xie, Y.}, \bibinfo{author}{Zhang, J.}, \bibinfo{author}{Xia, Y.}, \bibinfo{author}{Wu, Q.}, \bibinfo{year}{2022}a.
\newblock \bibinfo{title}{Unimiss: Universal medical self-supervised learning via breaking dimensionality barrier}, in: \bibinfo{booktitle}{European Conference on Computer Vision}, \bibinfo{organization}{Springer}. pp. \bibinfo{pages}{558--575}.
\bibitem[{Xie et~al.(2022b)Xie, Zhang, Cao, Lin, Bao, Yao, Dai and Hu}]{simmim}
\bibinfo{author}{Xie, Z.}, \bibinfo{author}{Zhang, Z.}, \bibinfo{author}{Cao, Y.}, \bibinfo{author}{Lin, Y.}, \bibinfo{author}{Bao, J.}, \bibinfo{author}{Yao, Z.}, \bibinfo{author}{Dai, Q.}, \bibinfo{author}{Hu, H.}, \bibinfo{year}{2022}b.
\newblock \bibinfo{title}{Simmim: A simple framework for masked image modeling}, in: \bibinfo{booktitle}{Proceedings of the IEEE/CVF conference on computer vision and pattern recognition}, pp. \bibinfo{pages}{9653--9663}.
\bibitem[{Yang et~al.(2022)Yang, Li, Tang, Zhu, Wang, Chen, Bai, Zhao and Ouyang}]{ratio2}
\bibinfo{author}{Yang, H.}, \bibinfo{author}{Li, X.}, \bibinfo{author}{Tang, S.}, \bibinfo{author}{Zhu, F.}, \bibinfo{author}{Wang, Y.}, \bibinfo{author}{Chen, M.}, \bibinfo{author}{Bai, L.}, \bibinfo{author}{Zhao, R.}, \bibinfo{author}{Ouyang, W.}, \bibinfo{year}{2022}.
\newblock \bibinfo{title}{Cycle-consistent masked autoencoder for unsupervised domain generalization}, in: \bibinfo{booktitle}{The Eleventh International Conference on Learning Representations}.
\bibitem[{Zhang et~al.(2023)Zhang, Zheng and Gu}]{sslmia}
\bibinfo{author}{Zhang, C.}, \bibinfo{author}{Zheng, H.}, \bibinfo{author}{Gu, Y.}, \bibinfo{year}{2023}.
\newblock \bibinfo{title}{Dive into the details of self-supervised learning for medical image analysis}.
\newblock \bibinfo{journal}{Medical Image Analysis} \bibinfo{volume}{89}, \bibinfo{pages}{102879}.
\bibitem[{Zhang et~al.(2022a)Zhang, Wang and Wang}]{lowlevel}
\bibinfo{author}{Zhang, Q.}, \bibinfo{author}{Wang, Y.}, \bibinfo{author}{Wang, Y.}, \bibinfo{year}{2022}a.
\newblock \bibinfo{title}{How mask matters: Towards theoretical understandings of masked autoencoders}.
\newblock \bibinfo{journal}{Advances in Neural Information Processing Systems} \bibinfo{volume}{35}, \bibinfo{pages}{27127--27139}.
\bibitem[{Zhang et~al.(2022b)Zhang, Wang and Wang}]{svd}
\bibinfo{author}{Zhang, Q.}, \bibinfo{author}{Wang, Y.}, \bibinfo{author}{Wang, Y.}, \bibinfo{year}{2022}b.
\newblock \bibinfo{title}{How mask matters: Towards theoretical understandings of masked autoencoders}.
\newblock \bibinfo{journal}{Advances in Neural Information Processing Systems} \bibinfo{volume}{35}, \bibinfo{pages}{27127--27139}.
\bibitem[{Zhang et~al.(2022c)Zhang, Wang and Wang}]{theoretical}
\bibinfo{author}{Zhang, Q.}, \bibinfo{author}{Wang, Y.}, \bibinfo{author}{Wang, Y.}, \bibinfo{year}{2022}c.
\newblock \bibinfo{title}{How mask matters: Towards theoretical understandings of masked autoencoders}.
\newblock \bibinfo{journal}{Advances in Neural Information Processing Systems} \bibinfo{volume}{35}, \bibinfo{pages}{27127--27139}.
\bibitem[{Zhang et~al.()Zhang, Tan, Yang, Huang and Yuan}]{er3}
\bibinfo{author}{Zhang, Y.}, \bibinfo{author}{Tan, Z.}, \bibinfo{author}{Yang, J.}, \bibinfo{author}{Huang, W.}, \bibinfo{author}{Yuan, Y.}, .
\newblock \bibinfo{title}{Matrix information theory for self-supervised learning}, in: \bibinfo{booktitle}{Forty-first International Conference on Machine Learning}.
\bibitem[{Zhou et~al.(2023a)Zhou, Lu, Chen, Yang and Yu}]{prlv2}
\bibinfo{author}{Zhou, H.Y.}, \bibinfo{author}{Lu, C.}, \bibinfo{author}{Chen, C.}, \bibinfo{author}{Yang, S.}, \bibinfo{author}{Yu, Y.}, \bibinfo{year}{2023}a.
\newblock \bibinfo{title}{A unified visual information preservation framework for self-supervised pre-training in medical image analysis}.
\newblock \bibinfo{journal}{IEEE Transactions on Pattern Analysis and Machine Intelligence} \bibinfo{volume}{45}, \bibinfo{pages}{8020--8035}.
\bibitem[{Zhou et~al.(2023b)Zhou, Liu, Bae, He, Samaras and Prasanna}]{mae3d}
\bibinfo{author}{Zhou, L.}, \bibinfo{author}{Liu, H.}, \bibinfo{author}{Bae, J.}, \bibinfo{author}{He, J.}, \bibinfo{author}{Samaras, D.}, \bibinfo{author}{Prasanna, P.}, \bibinfo{year}{2023}b.
\newblock \bibinfo{title}{Self pre-training with masked autoencoders for medical image classification and segmentation}, in: \bibinfo{booktitle}{2023 IEEE 20th International Symposium on Biomedical Imaging (ISBI)}, \bibinfo{organization}{IEEE}. pp. \bibinfo{pages}{1--6}.
\bibitem[{Zhou et~al.(2023c)Zhou, Greenspan and Shen}]{deepmia}
\bibinfo{author}{Zhou, S.K.}, \bibinfo{author}{Greenspan, H.}, \bibinfo{author}{Shen, D.}, \bibinfo{year}{2023}c.
\newblock \bibinfo{title}{Deep learning for medical image analysis}.
\newblock \bibinfo{publisher}{Academic Press}.
\bibitem[{Zhou et~al.(2021)Zhou, Sodha, Pang, Gotway and Liang}]{mg}
\bibinfo{author}{Zhou, Z.}, \bibinfo{author}{Sodha, V.}, \bibinfo{author}{Pang, J.}, \bibinfo{author}{Gotway, M.B.}, \bibinfo{author}{Liang, J.}, \bibinfo{year}{2021}.
\newblock \bibinfo{title}{Models genesis}.
\newblock \bibinfo{journal}{Medical image analysis} \bibinfo{volume}{67}, \bibinfo{pages}{101840}.

\end{thebibliography}

\renewcommand\thefigure{S\arabic{figure}} 
\setcounter{figure}{0} 
\renewcommand\thetable{S\arabic{table}} 
\setcounter{table}{0}

\cleardoublepage

\section*{Appendix}

\section{Datasets}
\noindent\textbf{Pre-training and downstream datasets.} We utilize thirteen datasets (a total of 9,995 CT scans) for pre-training, including BTCV~\citep{btcv}, Sliver07~\citep{sliver07}, CT-ORG~\citep{ctorg}, FLARE'22~\citep{flare}, CHAOS~\citep{chaos}, NaH-Seg~\citep{han}, KiPA22~\citep{kipa1, kipa2, kipa3, kipa4}, COVID-19~\citep{convid}, Pancreas-CT~\citep{pan}, LiTS~\citep{lits}, AbdomenCT-1k~\citep{abd1k}, LUNA16~\citep{luna}, and AbdomenAtlasMini 1.0~\citep{atlas10}. Details of the fine-tuning datasets used for downstream tasks are provided in Table~\ref{tab:split}. Specifically, we fine-tune seven datasets, including BTCV~\citep{btcv}, Sliver07~\citep{sliver07}, CT-ORG~\citep{ctorg}, FLARE'22~\citep{flare}, WORD~\citep{word}, AMOS~\citep{amos}, and BraTS 2021~\citep{brats21}, for one-shot and varying dataset proportions fine-tuning.

\begin{table}[h]
\vspace{-2mm}
\caption{Overview of downstream dataset.\label{tab:split}}
\vspace{-2mm}
\centering
\resizebox{0.9\linewidth}{!}
{
\begin{tabular}{l c r r r r r r}
\Xhline{1px} 
\multirow{2}{*}{Dataset} & \multirow{2}{*}{Modality} & \multicolumn{5}{c}{Train} & \multirow{2}{*}{Vaild} \\
\cline{3-7}
& & full & one-shot & 1\% & 10\% & 100\% (full) & \\
\hline
BTCV & CT & 24 & 1 & 1 & 2 & 24 & 6 \\
FLARE'22 & CT & 50 &  1 & 1 & 10 & 50 & 50 \\  
AMOS & CT/MRI & 240 & 1 & 2 & 25 & 240 & 120 \\
WORD & CT & 100 & 1 & - & - & - &  20  \\
Sliver07 & CT & 16 & 1 & - & - & - & 4  \\
CT-ORG & CT & 112 & 1 & - & - & - & 28 \\ 
BraTS 21 & MRI & 1000 & 1 & - & - & - & 251 \\
\Xhline{1px}
\end{tabular}
}
\vspace{-2mm}
\end{table}

\noindent\textbf{BTCV dataset.} The Multi-Atlas Labeling Beyond The Cranial Vault (BTCV) dataset~\citep{btcv} consists of 30 abdominal CT scans annotated at the pixel level for 13 organs by interpreters under the supervision of clinical radiologists at Vanderbilt University Medical Center. Following the previous works~\citep{unetr, swinunetr, voco, mim}, we utilize the same dataset split, \textit{e.g.}, with 24 scans for training and 6 scans for validation, for fair comparison. Additionally, we conduct random training dataset splits with one-shot, 1\%, and 10\% proportion (as shown in Table~\ref{tab:split}).

\noindent\textbf{AMOS dataset.} The Multi-Modality Abdominal Multi-Organ Segmentation Challenge (AMOS) dataset~\citep{amos} comprises 360 CT and MRI scans annotated for 15 abdominal organs. We adhere to the official split, using 240 samples for training and 120 samples for validation. We conduct random training dataset splits with different settings for one-shot, 1\%, and 10\% proportion (as shown in Table~\ref{tab:split}).

\begin{table}[!t]

\centering
\caption{Overview of pre-training and fine-tuning settings.\label{tab:settings}}
\vspace{-2mm}
\resizebox{0.9\linewidth}{!}
{
\begin{tabular}{l l}
\Xhline{1px} 
\multicolumn{1}{c}{\textit{{\color{Gray} Pre-training pre-processing}}} & \\
\hline
Spacing & $1.5 \times 1.5 \times 1.5~(mm)$ \\
Intensity & $[-175, 250]$ \\
Sub-volume size & $96 \times 96 \times 96$ \\
Sub-crops & 8 \\
Augmentation & Random Rotate, Flip, Scale, Shift \\
\Xhline{1px}
\multicolumn{1}{c}{\textit{{\color{Gray} Pre-training settings}}} & \\
\hline
Pre-training steps & 400K \\
Optimizer & AdamW \\
Weight decay & 0.05 \\
Optimizer momentum & $\beta_{1}, \beta_{2} = 0.9, 0.95$ \\ 
Optimizer LR & 1e-4 \\
Batch size & $24 \times 8 = 192$ \\
LR schedule & warmup cosine \\
Warm-up steps & 4K \\
\Xhline{1px}
\multicolumn{1}{c}{\textit{{\color{Gray} Fine-tuning pre-processing}}} & \\
\hline
Spacing & $1.5 \times 1.5 \times 1.5~(mm)$ \\
Intensity & $[-175, 250]$ \\
Sub-volume size & $96 \times 96 \times 96$ \\
Sub-crops & 4 \\
Augmentation & Random Rotate, Flip, Scale, Shift \\
\Xhline{1px}
\multicolumn{1}{c}{\textit{{\color{Gray} Fine-tuning settings}}} & \\
\hline
Optimizer & AdamW \\
Optimizer LR & 1e-4 \\
Weight decay & 1e-5 \\
Batch size & $1 \times 4 = 4$ \\
Swin batch size & 1 \\
Inference & sliding window \\
\Xhline{1px}
\end{tabular}
}

\vspace{-4mm}
\end{table}

\noindent\textbf{FLARE'22 dataset.} The FLARE'22 dataset, from the MICCAI 2022 Fast and Low-resource Semi-Supervised Abdominal Organ Segmentation Challenge~\citep{flare}, includes 100 annotated CT scans for the segmentation of 13 abdominal organs. We use the official training set of 50 CT scans and validation set of another 50 CT scans from different medical centers~\citep{flare_center}. We conduct random training dataset splits with different settings for one-shot, 1\%, 10\% proportion (as shown in Table~\ref{tab:split}).

\noindent\textbf{WORD dataset.} The large-scale whole abdominal organ dataset (WORD) dataset~\citep{word} contains 150 high-resolution CT scans, with 16 pixel-level organ annotations. We used the official training set of 100 CT scans and 20 CT scans for validation~\citep{word}. The WORD dataset is only utilized for the one-shot downstream task (as shown in Table~\ref{tab:split}).

\noindent\textbf{Sliver07 dataset.} The Sliver07 dataset~\citep{sliver07}, from the Segmentation of Liver Competition held in MICCAI 2007, comprises 20 CT scans for liver segmentation. We adopt an 80/20 train-validation split for the one-shot downstream task (as shown in Table~\ref{tab:split}).

\noindent\textbf{CT-ORG dataset.} The CT volumes with multiple organ segmentations (CT-ORG) dataset~\citep{ctorg} consists of 150 CT scans for the 5 organ and bone annotations from several clinical sites. We adopt an 80/20 train-validation split for the one-shot downstream task (as shown in Table~\ref{tab:split}).

\begin{table*}[t]

\centering
\caption{Hi-End-MAE variants.\label{tab:variant}}
\vspace{-2mm}
\resizebox{0.85\linewidth}{!}
{
\begin{tabular}{c | c | c | c }
\Xhline{1px} 

\multicolumn{4}{c}{Encoder} \\
\hline
variants &  ViT-B/16$^{(768)}$ & ViT-B/16$^{(1536)}$ & ViT-B/12$^{(1536)}$ \\

\hline
 
\multicolumn{1}{c|}{-} &

$\begin{bmatrix} 
    \text{SelfAttn, p. sz.}~16\text{,}
    \\
    \text{dim}~768,\text{head}~16
\end{bmatrix} \times 12$ & 

$\begin{bmatrix}
    \text{SelfAttn, p. sz.}~16\text{,}
    \\
    \text{dim}~1536,\text{head}~16
\end{bmatrix} \times 12$ & 

$\begin{bmatrix}
    \text{SelfAttn, p. sz.}~12\text{,}
    \\
    \text{dim}~1536,\text{head}~16
\end{bmatrix} \times 12$ \\ 

\hline
\multicolumn{4}{c}{Decoder} \\
\hline

 - & $\begin{bmatrix} 
    \text{SelfAttn, p. sz.}~16\text{,}
    \\
    \text{dim}~384\text{, head}~16
\end{bmatrix} \times 2$ & 

$\begin{bmatrix}
    \text{SelfAttn, p. sz.}~16\text{,}
    \\
    \text{dim}~528\text{, head}~16
\end{bmatrix} \times 2$ & 

$\begin{bmatrix}
    \text{SelfAttn, p. sz.}~12\text{,}
    \\
    \text{dim}~528\text{, head}~16
\end{bmatrix} \times 2$ \\

stage1 & $\begin{bmatrix} 
    \text{CrossAttn, p. sz.}~16\text{,}
    \\
    \text{dim}~384\text{, head}~16
\end{bmatrix} \times 1$ & 

$\begin{bmatrix}
    \text{CrossAttn, p. sz.}~16\text{,}
    \\
    \text{dim}~528\text{, head}~16
\end{bmatrix} \times 1$ & 

$\begin{bmatrix}
    \text{CrossAttn, p. sz.}~12\text{,}
    \\
    \text{dim}~528\text{, head}~16
\end{bmatrix} \times 1$ \\

stage2 & $\begin{bmatrix} 
    \text{CrossAttn, p. sz.}~16\text{,}
    \\
    \text{dim}~384\text{, head}~16
\end{bmatrix} \times 1$ & 

$\begin{bmatrix}
    \text{CrossAttn, p. sz.}~16\text{,}
    \\
    \text{dim}~528\text{,}\text{head}~16
\end{bmatrix} \times 1$ & 

$\begin{bmatrix}
    \text{CrossAttn, p. sz.}~12\text{,}
    \\
    \text{dim}~528\text{, head}~16
\end{bmatrix} \times 1$ \\

 stage3 & $\begin{bmatrix} 
    \text{CrossAttn, p. sz.}~16\text{,}
    \\
    \text{dim}~384\text{, head}~16
\end{bmatrix} \times 1$ & 

$\begin{bmatrix}
    \text{CrossAttn, p. sz.}~16\text{,}
    \\
    \text{dim}~528\text{, head}~16\text{,}
\end{bmatrix} \times 1$ & 

$\begin{bmatrix}
    \text{CrossAttn, p. sz.}~12\text{,}
    \\
    \text{dim}~528\text{, head}~16
\end{bmatrix} \times 1$ \\

\Xhline{1px}
\end{tabular}
} 

\vspace{-4mm}
\end{table*}

\begin{figure*}[t!]
    \centering
    \vspace{4mm}
    \includegraphics[width=0.99\textwidth]{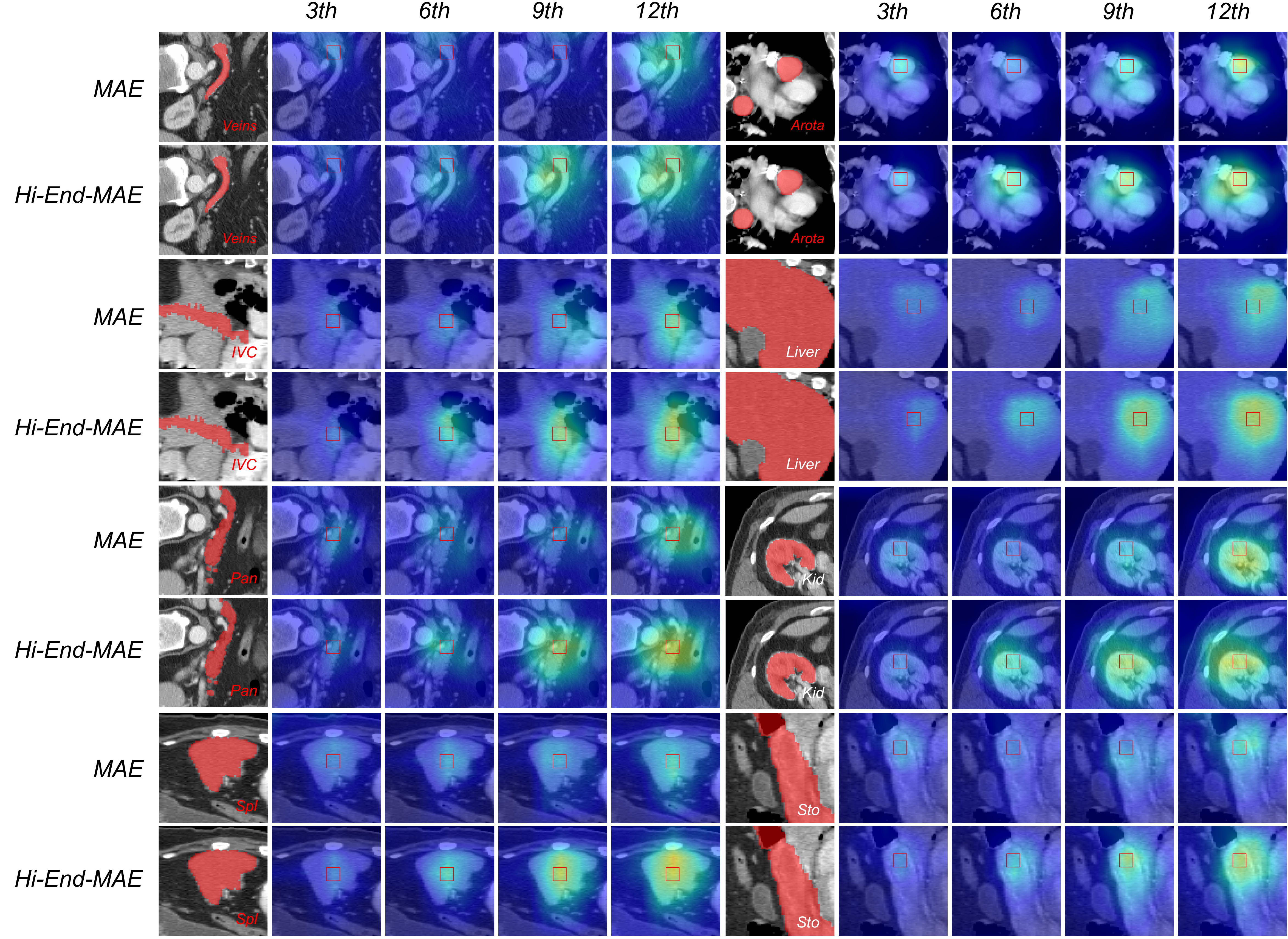}
    \vspace{-2mm}
    \caption{Visualization of attention maps in the 3th, 6th, 9th, and 12th layers of ViT-B/12$^{(1536)}$ for query patches (red box) on different organs, pre-trained by MAE and Hi-End-MAE. The attention maps are averaged across all attention heads. The abbreviations IVC, Pan, Kid, Spl, and Sto correspond to Inferior Vena Cava, Pancreas, Kidney, Spleen, and Stomach, respectively.}
    \label{fig:avg_attn_map}
\end{figure*}

\begin{figure*}[t]
    \centering
    \includegraphics[width=0.99\textwidth]{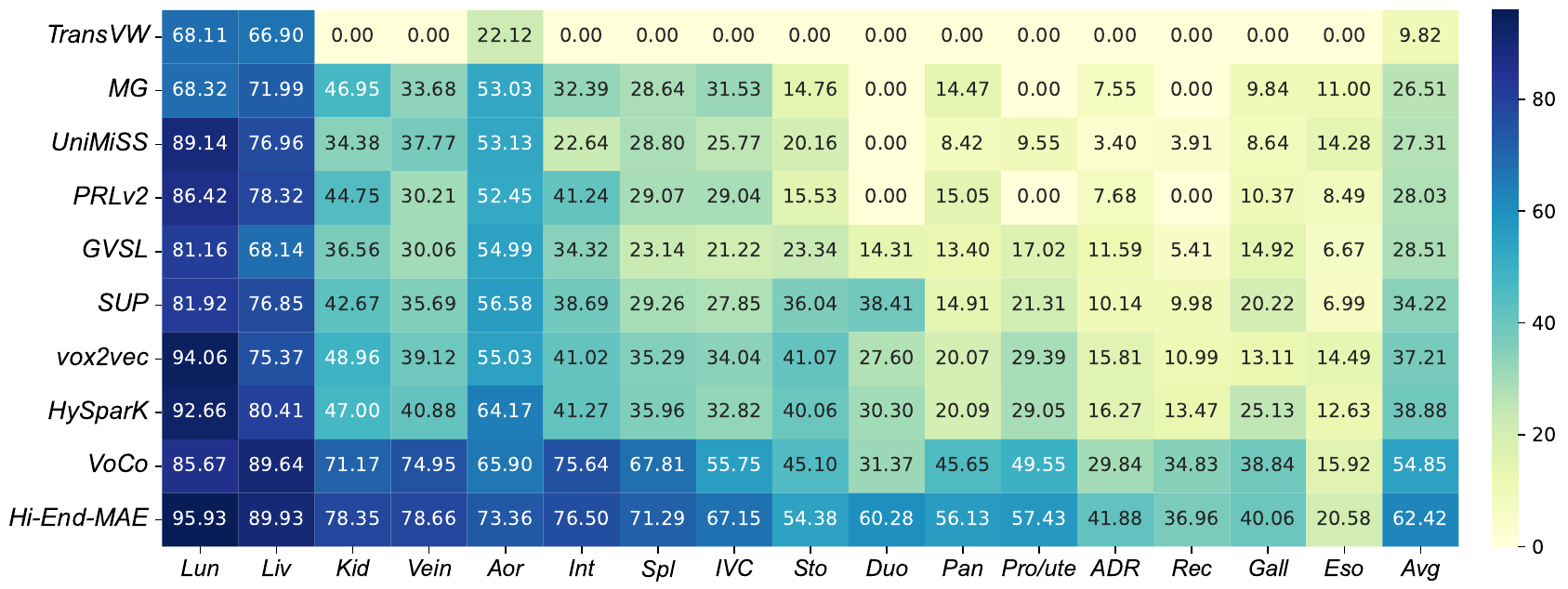}
    \vspace{-2mm}
    \caption{Comparative analysis one-shot segmentation results across 16 targets in terms of DSC (\%) performance.}
    \label{fig:total_oneshot}
    \vspace{-2mm}
\end{figure*}

\begin{figure*}[t]
    \centering
    \includegraphics[width=0.99\textwidth]{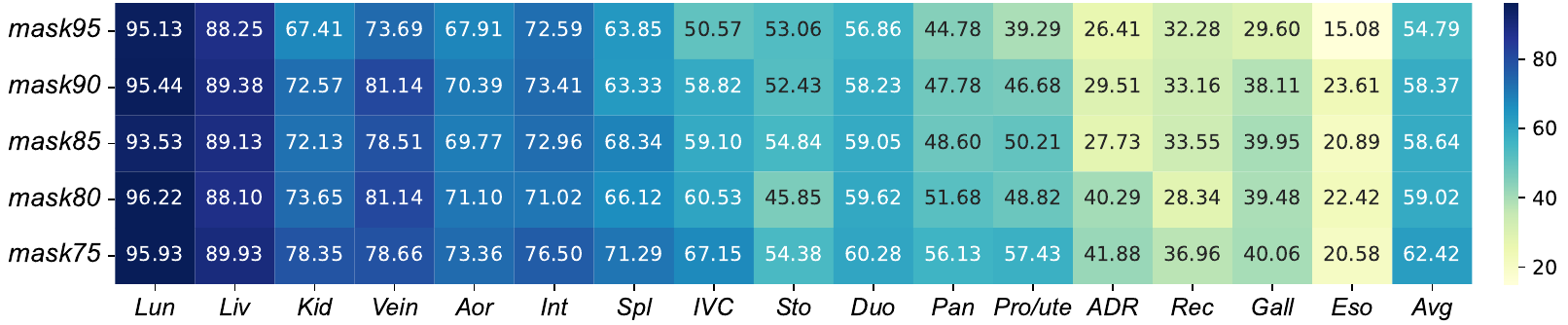}
    \vspace{-2mm}
    \caption{Comparative analysis of different mask ratios results across 16 segmentation targets in terms of DSC (\%) performance.}
    \vspace{-2mm}
    \label{fig:total_mask_ratio}
\end{figure*}

\begin{figure*}[!t]
    \centering
    \includegraphics[width=0.99\textwidth]{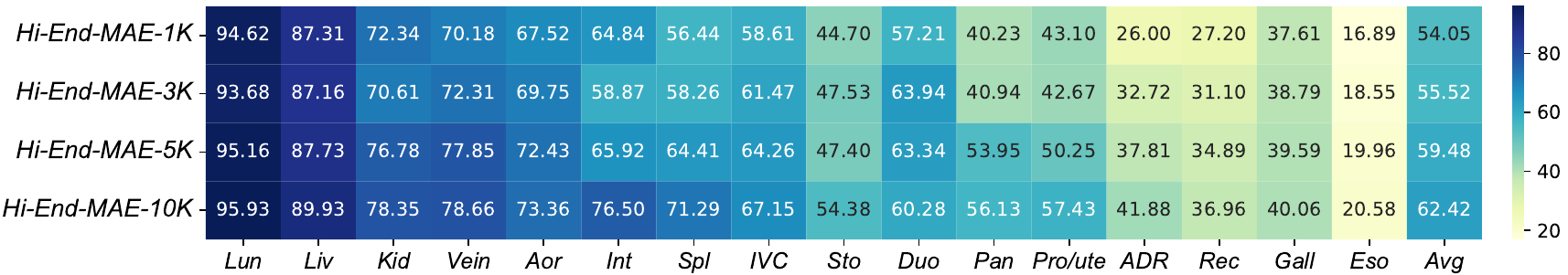}
    \vspace{-2mm}
    \caption{Comparative analysis of different data scale results across 16 segmentation targets in terms of DSC (\%) performance.}
    \vspace{-2mm}
    \label{fig:total_data_percent}
\end{figure*}

\noindent\textbf{BraTS 21 dataset.} The BraTS 21 dataset~\citep{brats21}, from the BraTS 2021 challenge of brain tumors by providing 1251 MRI scans with pixel-level annotations. Following the previous works~\citep{swinunetr, voco}, we utilize the same dataset split, \textit{e.g.}, with 1000 scans for training and 251 scans for validation, for fair comparison. In this paper, we evaluate the ability of model generalization on the BraTs 21 dataset through one-shot fine-tuning.

\section{Implementation Details}
\noindent\textbf{Pre-training and fine-tuning settings.} The details of our pre-training and fine-tuning settings are shown in Table~\ref{tab:settings}. For pre-training task, We sample the pre-training sub-volumes of $96 \times 96 \times 96$ voxels by ratios of positive and negative as 3:1 in 8 sub-crops. Augmentation probabilities for random flip, rotation, intensities scaling, and shifting are set to 0.5, 0.3, 0.1, 0.1, respectively. For downstream task, the sample ratios of positive and negative are as 1:1 in 4 sub-crops. Augmentation probabilities for random flip, rotation, intensities scaling, and shifting are set to 0.2, 0.2, 0.1, 0.1, respectively. The training epochs for fine-tuning are set to 5000, 5000, 2000, 1000 for one-shot, 1\%, 10\%, 100\% proportion downstream segmentation tasks, respectively.

\noindent\textbf{Hi-End-MAE variants.} The detailed Hi-End-MAE variants are shown in Table~\ref{tab:variant}. ``SelfAttn" and ``CrossAttn" denote the self attention block and cross attention block, respectively. ``p. sz. 12" indicates the patch size of $12 \times 12 \times 12$.

\begin{figure*}[t]
    \centering
    \includegraphics[width=0.99\textwidth]{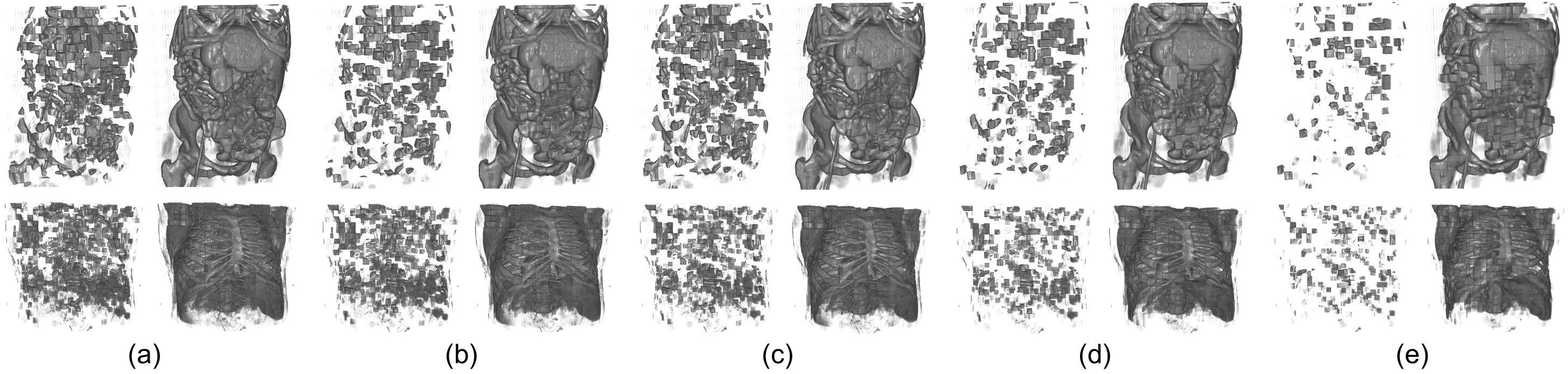}
    \vspace{-2mm}
    \caption{Abdomen (row 1) and Lung (row 2) CT reconstruction results of Hi-End-MAE with different mask ratios: (a) mask 80\% (b) mask 85\% (c) mask 85\% (d) mask 90\% and (e) mask 95\%. For each double, we show the masked image (left), and our Hi-End-MAE reconstruction result (right).}
    \vspace{-2mm}
    \label{fig:recon}
\end{figure*}

\begin{figure*}[t]
    \centering
    \includegraphics[width=0.92\textwidth]{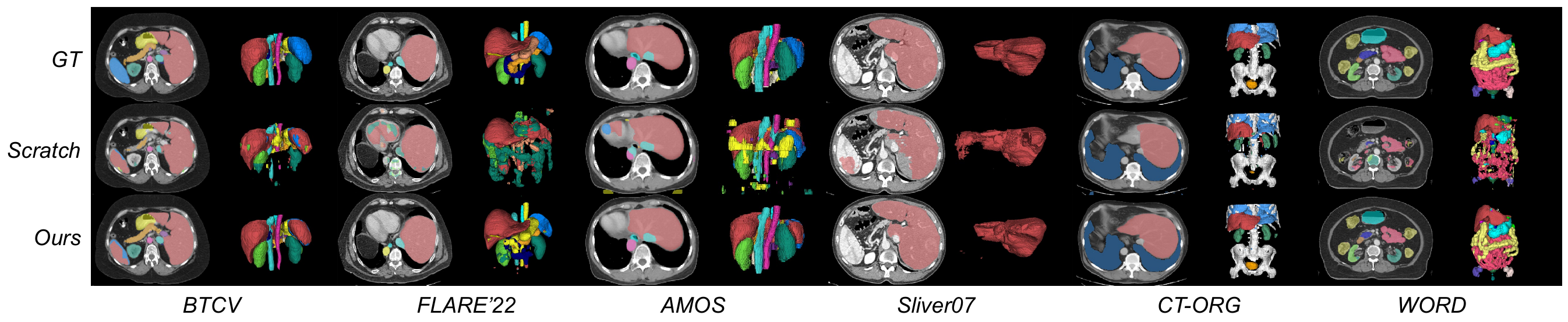}
    \vspace{-2mm}
    \caption{Qualitative visualization of one-shot segmentation results on BTCV~\citep{btcv}, FLARE’22~\citep{flare}, AMOS~\citep{amos}, Sliver07~\citep{sliver07}, CT-ORG~\citep{ctorg}, and  WORD~\citep{word}. Left/Right: Slice/Volume segmentation results on the same dataset.}
    \label{fig:vis}
    \vspace{-2mm}
\end{figure*}

\section{Analysis}
\noindent\textbf{Local attention patterns.} We visualize the slice-level multi-head average attention map from different ViT layers. As shown in Fig.\ref{fig:avg_attn_map}, compared to MAE (decoder-driven reconstruction), our Hi-End-MAE benefits from a more refined encoder-driven dense decoding mechanism that learns localized anatomical representations with stronger perceptual capabilities, which could gain the ability to model complex relationships representation.

\section{Experiments}
\noindent\textbf{Target-specific analysis for one-shot medical segmentation.}  We present the average one-shot segmentation semantic results across six datasets, \textit{e.g.}, BTCV~\citep{btcv}, CT-ORG~\citep{ctorg}, Sliver07~\citep{sliver07}, WORD~\citep{word}, AMOS~\citep{amos}, and FLARE’22~\citep{flare}. The overall results are illustrated in Table~\ref{fig:total_oneshot}. Hi-End-MAE delivers outstanding one-shot segmentation performance across segmentation targets of varying scales, \textit{e.g.}, achieving a notable improvement in kidney with 7.18\&, Arota with 7.46\%, Inferior Vena Cava with 11.4\%, Stomach with 9.28\%, Duodenum with 21.87\%, Pancreas with 10.48\%, Prostate/Uterus with 7.88\%, and Adrenal Gland with 12.04\%.

\noindent\textbf{Comparison across various data proportions.} We fine-tune pre-trained models on BTCV~\citep{btcv}, AMOS~\citep{amos} and FLARE'22~\citep{flare} using 1\%, 10\%, and 100\% proportion of the dataset. The DSC scores of different segmentation targets are shown in Table~\ref{tab:btcv}, Table~\ref{tab:amos}, and Table~\ref{tab:flare}, respectively. Our method demonstrates a significant advantage in downstream tasks, particularly in scenarios with limited annotation.

\noindent\textbf{Target-specific analysis for mask ratios and data scales.} We present the average one-shot segmentation semantic results across varying mask ratios and data scales on six datasets: BTCV~\citep{btcv}, CT-ORG~\citep{ctorg}, Sliver07~\citep{sliver07}, WORD~\citep{word}, AMOS~\citep{amos}, and FLARE’22~\citep{flare}. The specific segmentation target results on different mask ratios and data scales are shown in Fig.\ref{fig:total_mask_ratio} and Fig.\ref{fig:total_data_percent}, respectively.

\renewcommand{\multirowsetup}{\centering}  
\begin{table*}[t]
\centering
\caption{Comparison of different methods with 1\%, 10\%, and 100\% proportions on BTCV~\citep{btcv}. We report the DSC (\%) performance. \textbf{val} (bold) / \underline{val} (underline) : top method / second method. $\dagger$ denotes we utilize official pre-training weights. $\ddagger$ denotes the results are copied from~\citep{voco}. The abbreviations Spl, RKid, LKid, Gall, Eso, Liv, Sto, Aor, IVC, Veins, Pan, RAG, and LAG correspond to Spleen, Right kidney, Left kidney, Gallbladder, Esophagus, Liver, Stomach, Aorta, Inferior Vena Cava, Veins, Pancreas, Right Adrenal Gland, and Left Adrenal Gland, respectively.}
\vspace{-2mm}
\resizebox{0.99\linewidth}{!}
{
\begin{tabular}{l c c c c c c c c c c c c c c}
\Xhline{1px} 
\multirow{2}{*}{Method} & \multicolumn{13}{c}{BTCV (100\%)}  & \multirow{2}{*}{Avg} \\
\cline{2-14}
& Spl & RKid & LKid & Gall & Eso & Liv & Sto & Aor & IVC & Veins & Pan & RAG & LAG & \\
\hline
\multicolumn{1}{c}{\textit{{\color{Gray} From scratch}}} \\
UNETR$\dagger$$\ddagger$ & 93.02 & 94.13 & 94.12 & 66.99 & 70.87 & 96.11 & 77.27 & 89.22 & 82.10 & 70.16 & 76.65 & 65.32 & 59.21 & 79.82 \\
SwinUNETR$\dagger$$\ddagger$ & 94.06 & 93.54 & 93.80 & 65.51 & 74.60 & 97.09 & 75.94 & \textbf{91.80} & 82.36 & 73.63 & 75.19 & 68.00 & 61.11 & 80.53 \\
\hline
\multicolumn{1}{c}{\textit{{\color{Gray} Medical SSL}}} \\
MG$\dagger$$\ddagger$ & 91.99 & 93.52 & 91.81 & 65.11 & \textbf{76.14} & 95.98 & \underline{86.88} &  89.29 & 83.59 & 71.79 & 81.62 & 67.97 & 63.18 & 81.45 \\
PRLv2$\dagger$$\ddagger$ & 95.50 & 91.43 & 89.52 & \textbf{76.15} & 73.54 & \underline{97.28} & 79.64 & 90.16 & 84.17 & 75.20 & 78.71 & 68.74 & 62.93 & 81.74 \\
SUP$\dagger$$\ddagger$ & 95.25 & 93.16 & 92.97 & 63.62 & 73.96 & 96.21 & 79.32 & 89.98 & 83.19 & 76.11 & \underline{82.25} & 68.99 & 65.11 & 81.54 \\
GVSL$\dagger$$\ddagger$ & 95.27 & 91.22 & 92.25 & 72.69 & 73.56 & 96.44 & 82.40 & 88.90 & 84.22 & 70.84 & 76.42 & 67.48 & 63.25 & 81.87 \\
VoCo$\dagger$$\ddagger$ & \underline{95.73} & \textbf{96.53} & \underline{94.48} & \underline{76.02} & 75.60 & \textbf{97.41} & 78.43 & \underline{91.21} & \underline{86.12} & \textbf{78.19} & 80.88 & \textbf{71.47} & \underline{67.88} & \underline{83.85} \\
\rowcolor{gray!15} \textbf{Hi-End-MAE}  & \textbf{96.33} & \underline{94.87} & \textbf{94.87} & 64.55 & \underline{75.72} & 97.16 & \textbf{91.80} & 90.64 & \textbf{86.91} & \underline{78.16} & \textbf{85.68} & \underline{70.44} & \textbf{71.79} & \textbf{84.53} \\
\Xhline{1px} 

& \multicolumn{13}{c}{BTCV (10\%)} & \\
\hline
\multicolumn{1}{c}{\textit{{\color{Gray} From scratch}}} \\
UNETR & 60.25 & 76.85 & 64.96 & 30.47 & 41.77 & 86.73 & 22.01 & 63.15 & 48.58 & 39.90 & 22.28 &0.00&0.00& 42.85 \\
SwinUNETR & 81.59 & 80.86 & 79.78 & 36.48 & 37.55 & 88.96 & 25.41 & 57.90 & 57.99 & 49.20 & 12.05 & 43.98 & 15.49 & 51.33 \\
\hline
\multicolumn{1}{c}{\textit{{\color{Gray} General SSL}}} \\
SparK & 69.60 & 60.61 & 55.83 & 35.04 & 26.46 & 88.47 & 37.67 & 67.96 & 60.59 & 52.86 & 30.73 & 49.42 & 31.11 & 51.26 \\
MAE & 86.87 & 92.64 & 92.75 & 47.09 & 60.17 & 95.00 & 71.69 & 86.37 & \underline{81.96} & \underline{68.51} & \textbf{74.23} & 59.96 & 57.81 & 75.01 \\
\hline
\multicolumn{1}{c}{\textit{{\color{Gray} Medical SSL}}} \\
MG$\dagger$ & 59.10 & 62.29 & 52.00 & 24.28 &0.00& 82.92 & 36.49 & 69.95 & 62.16 & 45.13 & 0.143 &0.00&0.00& 38.04 \\
TransVW$\dagger$ &0.00&0.00&0.00&0.00&0.00& 81.23 & 28.21 &0.00&0.00&0.00&0.00&0.00&0.00& 8.42 \\
UniMiSS$\dagger$ & 48.47 & 74.86 & 72.64 & 25.82 & 30.56 & 88.62 & 28.38 & 59.93 & 56.67 & 53.93 & 39.00 & 33.15 &0.00& 47.08 \\
SUP & 64.41 & 76.44 & 70.29 & 30.90 & 33.36 & 86.85 & 24.78 & 64.86 & 57.66 & 54.66 & 16.47 & 37.82 & 27.14 & 49.67 \\
PRLv2$\dagger$ & 71.98 & 62.80 & 46.66 &0.00&0.00& 88.37 & 19.14 & 57.99 & 49.34 &0.00&0.00&0.00&0.00& 30.48 \\
GVSL$\dagger$ & 52.82 & 33.54 & 34.05 & 35.44 & 41.98 & 90.91 & 33.14 & 74.80 & 44.95 & 34.78 & 18.86 & 25.83 & 22.06 & 41.79 \\
vox2vec$\dagger$ & 65.02 & 71.73 & 68.90 & 22.30 & 33.50 & 90.44 & 36.57 & 73.32 & 62.13 & 44.02 & 39.28 & 43.07 & 22.70 & 51.77 \\
HySparK$\dagger$ & 58.79 & 61.56 & 57.58 & 45.76 & 29.77 & 89.95 & 43.40 & 68.37 & 60.33 & 52.28 & 19.50 & 48.15 & 34.55 & 51.54 \\
VoCo$\dagger$ & \underline{91.73} & \underline{92.83} & \underline{93.18} & \textbf{65.13} & \textbf{67.60} & \underline{95.58} & \underline{78.76} & \underline{86.81} & 78.18 & 66.58 & 71.05 & \textbf{64.53} & \textbf{59.98} & \underline{77.85}  \\
\rowcolor{gray!15} \textbf{Hi-End-MAE} & \textbf{92.31} & \textbf{93.47} & \textbf{93.46} & \underline{63.56} & \underline{66.14} & \textbf{96.03} & \textbf{79.97} & \textbf{88.74} & \textbf{83.66} & \textbf{69.98} & \underline{71.84} & \underline{62.40} & \underline{59.64} & \textbf{78.56} \\
\Xhline{1px} 
& \multicolumn{13}{c}{BTCV (1\%)} & \\
\hline
\multicolumn{1}{c}{\textit{{\color{Gray} From scratch}}} \\
UNETR & 30.60 & 49.93 & 35.31 & 4.555 & 28.35 & 83.21 & 13.72 & 39.16 & 30.01 & 23.52 & 6.042 & 14.26 & 5.923 & 28.05 \\
SwinUNETR & 25.76 & 41.29 & 21.58 & 10.84 & 27.85 & 80.97 & 14.03 & 49.00 & 40.43 & 20.23 & 1.995 & 25.96 & 0.189 & 27.71 \\
\hline
\multicolumn{1}{c}{\textit{{\color{Gray} General SSL}}} \\
SparK & 41.65 & 41.08 & 40.09 & 9.036 & 25.79 & 81.29 & 15.17 & 52.37 & 38.45 & 19.47 & 3.298 & 19.64 & 11.62 & 30.69 \\
MAE & 71.68 & 87.19 & 87.68 & \underline{21.58} & 55.09 & 88.96 & 53.41 & 80.81 & \underline{73.27} & \underline{58.41} & \underline{49.00} & 47.39 & 32.02 & 62.04 \\
\hline
\multicolumn{1}{c}{\textit{{\color{Gray} Medical SSL}}} \\
MG$\dagger$ & 45.95 & 32.98 & 46.88 &0.00&0.00& 74.28 & 15.31 & 69.50 & 47.15 & 45.20 & 3.174 &0.00&0.00& 29.27 \\
TransVW$\dagger$ &0.00&0.00&0.00&0.00&0.00& 73.25 &0.00&0.00&0.00&0.00&0.00&0.00&0.00& 5.634 \\
UniMiSS$\dagger$ & 49.62 & 58.16 & 52.88 & 1.752 & 21.68 & 80.67 & 16.30 & 53.55 & 41.77 & 27.05 & 3.741 & 15.06 & 6.044 & 32.95 \\
SUP$\dagger$ & 20.14 & 45.41 & 37.65 & 8.281 & 29.73 & 81.66 & 13.50 & 51.83 & 35.21 & 20.28 & 0.098 & 26.72 & 3.141 & 28.74 \\
PRLv2$\dagger$ & 42.37 & 25.11 & 46.39 &0.00&0.00& 82.12 & 14.19 & 48.80 & 35.12 & 17.89 & 0.041 &0.00&0.00& 24.01 \\
GVSL$\dagger$ & 36.48 & 18.09 & 22.78 & 2.828 & 17.84 & 76.21 & 11.83 & 53.68 & 32.21 & 29.14 & 4.923 & 16.71 & 0.481 & 24.86 \\
vox2vec$\dagger$ & 52.08 & 30.36 & 37.30 & 10.92 & 43.86 & 86.65 & 14.72 & 66.05 & 47.59 & 27.99 & 5.075 & 32.91 & 3.176 & 35.29 \\
HySparK$\dagger$ & 36.77 & 38.86 & 48.45 & 17.83 & 41.83 & 79.53 & 17.91 & 63.49 & 39.71 & 34.71 & 1.399 & 25.31 & 19.66 & 35.81 \\
VoCo$\dagger$ & \underline{83.61} & \textbf{89.17} & \textbf{90.66} & 20.56 & \underline{60.38} & \underline{91.80} & \underline{61.94} & \underline{83.04} & 70.03 & 50.98 & 14.01 & \textbf{56.76} & \textbf{50.32} & \underline{63.33} \\
\rowcolor{gray!15} \textbf{Hi-End-MAE} & \textbf{85.55} & \underline{87.22} & \underline{89.00} & \textbf{24.43} & \textbf{67.78} & \textbf{93.48} & \textbf{72.22} & \textbf{84.03} & \textbf{74.28} & \textbf{64.02} & \textbf{61.83} & \underline{53.28} & \underline{47.51} & \textbf{69.59} \\

\Xhline{1px}
\end{tabular}
}

\vspace{-3mm}
\label{tab:btcv}
\end{table*}

\renewcommand{\multirowsetup}{\centering}  
\begin{table*}[t]
\centering
\caption{Comparison of different methods with 1\%, 10\%, and 100\% proportions on AMOS~\citep{amos}. We report the DSC (\%) performance. \textbf{val} (bold) / \underline{val} (underline) : top method / second method. $\dagger$ denotes we utilize official pre-training weights. The abbreviations Spl, RKid, LKid, Gall, Eso, Liv, Sto, Aor, IVC, Pos, Pan, RAG, LAG, Duo, Bla and Pro/Ute correspond to Spleen, Right kidney, Left kidney, Gallbladder, Esophagus, Liver, Stomach, Aorta, Postcava, Pancreas, Right Adrenal Gland, Left Adrenal Gland, Duodenum, Bladder, and Prostate/Uterus, respectively.}
\vspace{-2mm}
\resizebox{0.99\linewidth}{!}
{
\begin{tabular}{l c c c c c c c c c c c c c c c c}
\Xhline{1px} 
\multirow{2}{*}{Method} & \multicolumn{15}{c}{AMOS (100\%)}  & \multirow{2}{*}{Avg} \\
\cline{2-16}
& Spl & RKid & LKid & Gall & Eso & Liv & Sto & Aor & Pos & Pan & RAG & LAG & Duo & Bla & Pro/Ute & \\
\hline
\multicolumn{1}{c}{\textit{{\color{Gray} From scratch}}} \\
UNETR & 93.20 & 90.66 & 92.22 & 66.04 & 71.90 & 95.31 & 82.07 & 90.31 & 82.18 & 74.91 & 65.87 & 63.65 & 63.88 & 63.52 & 59.61 & 77.02 \\
SwinUNETR & 95.26 & 93.34 & 94.35 & 75.95 & 79.02 & 96.71 & 88.10 & 93.11 & 87.70 & 82.92 & 70.80 & 71.20 & 75.12 & 69.14 & 64.96 & 82.51 \\
\hline
\multicolumn{1}{c}{\textit{{\color{Gray} General SSL}}} \\
SparK & 95.64 & 93.71 & 95.23 & \underline{78.50} & 80.98 & 97.01 & 91.30 & 93.71 & 88.92 & 84.65 & 71.65 & 73.59 & 77.92 & 72.70 & 65.58 & 84.07 \\
MAE & 95.94 & 93.83 & 95.24 & 75.57 & 80.02 & 97.16 & 90.43 & 93.53 & 88.28 & 84.33 & 72.19 & 71.65 & 76.97 & 72.55 & 66.54 & 83.61 \\
\hline
\multicolumn{1}{c}{\textit{{\color{Gray} Medical SSL}}} \\
MG$\dagger$ & 93.84 & 91.51 & 91.92 & 75.30 &0.00& 93.82 & 88.44 & 93.24 & 88.22 & 83.62 &0.00&0.00& 74.56 & 70.26 &0.00& 62.99 \\
TransVW$\dagger$ & 94.01 & 92.12 & 92.19 & 74.31 & 80.36 & 95.37 & 88.71 & 93.23 & 88.35 & 83.65 & 72.18 & 72.71 & 76.07 & 70.46 & 64.88 & 82.58 \\
UniMiSS$\dagger$ & 94.19 & 92.40 & 93.83 & 71.76 & 72.73 & 96.09 & 86.63 & 91.43 & 84.63 & 79.16 & 66.52 & 67.27 & 70.48 & 68.90 & 62.64 & 79.92 \\
SUP$\dagger$ & 95.07 & 93.40 & 94.36 & 75.79 & 79.22 & 96.69 & 88.12 & 93.05 & 87.48 & 82.28 & 71.25 & 70.78 & 74.63 & 69.13 & 65.48 & 82.45 \\
PRLv2$\dagger$ & 89.20 & 87.69 & 88.14 & 67.83 &0.00& 93.97 & 81.07 & 91.24 & 82.02 & 71.64 &0.00&0.00& 59.28 &0.00&0.00& 54.14 \\
GVSL$\dagger$ & 94.22 & 92.46 & 93.26 & 73.95 & 78.63 & 96.17 & 87.84 & 92.36 & 86.23 & 81.35 & 70.25 & 69.97 & 73.11 & 67.91 & 62.99 & 81.38 \\
vox2vec$\dagger$ & 88.34 & 89.79 & 87.54 & 66.51 & 69.39 & 93.62 & 77.95 & 90.45 & 81.41 & 70.69 & 63.82 & 60.58 & 60.71 & 64.97 & 55.85 & 74.78 \\
HySparK$\dagger$ & \underline{96.08} & \textbf{94.43} & \underline{67.44} & \textbf{95.92} & \textbf{79.59} & \textbf{83.78} & \textbf{97.42} & \textbf{92.22} & \textbf{94.33} & \textbf{90.26} & \textbf{86.29} & \textbf{74.86} & \textbf{76.27} & \textbf{80.43} & \underline{74.27} & \textbf{85.58} \\
VoCo$\dagger$ & 95.57 & 94.07 & 95.07 & 78.42 & 81.77 & 96.89 & 90.66 & 93.93 & 89.25 & 85.37 & \underline{73.49} & 74.43 & 78.52 & 72.13 & 66.90 & 84.44 \\
\rowcolor{gray!15} \textbf{Hi-End-MAE} & \textbf{96.15} & \underline{94.09} & \textbf{68.43} & \underline{95.86} & 77.18 & \underline{82.55} & \underline{97.38} & \underline{92.17} & \underline{94.09} & \underline{89.47} & \underline{85.74} & 73.27 & \underline{74.51} & \underline{79.18} & \textbf{74.52} & \underline{84.98}  \\
\Xhline{1px} 
 & \multicolumn{15}{c}{AMOS (10\%)}  &  \\
\hline
\multicolumn{1}{c}{\textit{{\color{Gray} From scratch}}} \\
UNETR & 74.37 & 72.46 & 70.27 & 51.47 & 53.96 & 79.27 & 61.19 & 76.42 & 66.08 & 55.37 & 49.17 & 37.69 & 40.12 & 60.46 & 52.66 & 60.06 \\
SwinUNETR & 78.14 & 74.17 & 70.25 & 53.80 & 57.74 & 82.73 & 66.29 & 81.72 & 69.09 & 58.55 & 50.87 & 41.35 & 47.74 & 62.24 & 56.99 & 63.45 \\
\hline
\multicolumn{1}{c}{\textit{{\color{Gray} General SSL}}} \\
SparK & 86.05 & 86.31 & 84.81 & 59.63 & 63.71 & 89.94 & 75.60 & \textbf{88.59} & 76.68 & 67.63 & 60.43 & 53.77 & 60.26 & 63.60 & 58.14 & 71.68 \\
MAE & 82.38 & \underline{86.69} & 59.09 & 83.29 & 60.30 & 66.69 & 90.42 & \underline{77.64} & \underline{88.43} & 77.65 & 72.11 & \underline{61.61} & 60.23 & \underline{60.80} & \underline{66.79} & 72.94 \\
\hline
\multicolumn{1}{c}{\textit{{\color{Gray} Medical SSL}}} \\
MG$\dagger$ & 74.68 & 74.87 & 73.70 &0.00&0.00& 87.56 & 66.00 & 84.86 & 73.32 & 60.49 &0.00&0.00& 45.95 & 62.61 &0.00& 46.94 \\
TransVW$\dagger$ & 77.81 & 76.15 & 74.35 & 58.29 & 64.26 & 87.49 & 69.98 & 87.32 & 75.15 & 61.11 & 57.43 & 44.73 & 49.64 & 64.02 & 55.80 & 66.91 \\
UniMiSS$\dagger$ & 83.35 & 75.73 & 78.99 & 54.28 & 60.09 & 89.46 & 68.40 & 82.95 & 69.51 & 63.04 & 52.50 & 48.44 & 49.80 & 62.75 & 55.77 & 66.34 \\
SUP$\dagger$ & 77.34 & 78.32 & 73.15 & 53.70 & 60.62 & 82.99 & 65.53 & 83.08 & 70.97 & 60.16 & 53.72 & 46.89 & 49.74 & 62.27 & 55.73 & 64.95 \\
PRLv2$\dagger$  & 73.23 & 75.92 & 77.21 &0.00&0.00& 78.15 & 66.17 & 86.40 & 67.45 &0.00&0.00&0.00&0.00& 61.52 &0.00& 39.07 \\
GVSL$\dagger$  & 80.09 & 72.98 & 73.48 & 50.58 & 59.84 & 81.77 & 66.17 & 85.84 & 70.66 & 57.79 & 46.74 & 43.14 & 44.66 & 61.89 & 56.04 & 63.45 \\
vox2vec$\dagger$  & 75.61 & 73.14 & 73.80 & 52.68 & 58.05 & 78.53 & 67.46 & 83.21 & 66.13 & 57.35 & 45.79 & 38.10 & 47.64 & 62.37 & 54.61 & 62.30 \\
HySparK$\dagger$  & 75.71 & 80.04 & 74.62 & 55.63 & 57.68 & 82.39 & 68.85 & 82.49 & 72.80 & 56.21 & 42.61 & 45.64 & 51.13 & 62.69 & 56.23 & 64.32 \\
VoCo$\dagger$  & \textbf{89.35} & 81.64 & \underline{61.57} & \underline{85.30} & \underline{62.39} & \textbf{67.95} & \underline{90.60} & 76.09 & 87.83 & \textbf{80.07} & \underline{72.44} & 59.47 & \underline{60.38} & 60.64 & 64.27 & \underline{73.34} \\
\rowcolor{gray!15} \textbf{Hi-End-MAE}  & \underline{89.07} & \textbf{90.52} & \textbf{63.16} & \textbf{90.20} & \textbf{62.72} & \underline{67.54} & \textbf{93.26} & \textbf{78.30} & 87.19 & \underline{79.59} & \textbf{76.62} & \textbf{62.00} & \textbf{64.11} & \textbf{65.48} & \textbf{67.71} & \textbf{75.84} \\
\Xhline{1px} 
 & \multicolumn{15}{c}{AMOS (1\%)}  &  \\
\hline
\multicolumn{1}{c}{\textit{{\color{Gray} From scratch}}} \\
UNETR  & 28.83 & 21.01 & 19.53 & 26.21 & 11.86 & 72.55 & 35.04 & 33.54 & 30.53 & 8.181 & 5.494 & 1.366 & 4.098 & 33.62 & 23.16 & 23.67 \\
SwinUNETR  & 45.56 & 29.01 & 29.87 & 18.64 & 20.68 & 78.89 & 35.03 & 48.73 & 33.12 & 14.33 & 11.54 & 5.242 & 4.915 & 36.22 & 22.40 & 28.94 \\
\hline
\multicolumn{1}{c}{\textit{{\color{Gray} General SSL}}} \\
SparK  & 53.92 & 49.97 & 48.45 & 16.12 & 36.78 & 71.06 & 49.10 & 70.68 & 42.42 & 17.81 & 10.17 & 11.12 & 13.77 & 28.16 & 22.60 & 36.14 \\
MAE  & 67.62 & 72.90 & \underline{48.99} & 75.55 & 34.95 & 44.29 & \textbf{86.68} & 65.72 & 75.79 & 63.37 & 43.95 & \underline{28.19} & \textbf{44.66} & \underline{26.48} & 40.94 & 54.67 \\
\hline
\multicolumn{1}{c}{\textit{{\color{Gray} Medical SSL}}} \\
MG$\dagger$  & 37.00 & 30.65 & 30.78 & 20.77 &0.00& 75.23 & 35.12 & 59.42 & 39.17 & 6.713 &0.00&0.00& 3.910 & 31.48 & 15.48 & 25.72 \\
TransVW$\dagger$  & 20.27 & 26.30 & 26.51 &0.00&0.00& 71.11 & 32.09 & 54.82 & 32.16 & 8.989 &0.00& 1.640 & 4.202 &0.00& 2.670 & 18.72 \\
UniMiSS$\dagger$  & 53.58 & 34.17 & 36.66 & 21.38 & 16.00 & 82.24 & 38.27 & 38.96 & 36.36 & 12.62 & 3.501 & 0.811 & 4.045 & 35.64 & 28.08 & 29.49 \\
SUP$\dagger$  & 49.97 & 28.37 & 28.83 & 15.26 & 18.08 & 72.88 & 29.65 & 38.37 & 31.34 & 12.95 & 11.52 &0.00& 5.635 & 16.74 & 24.29 & 25.60 \\
PRLv2$\dagger$  & 41.59 & 26.29 & 26.86 &0.00&0.00& 70.50 & 35.64 & 56.01 & 34.51 & 3.044 &0.00&0.00& 1.375 & 20.20 &0.00& 21.07 \\
GVSL$\dagger$  & 42.21 & 20.83 & 24.83 & 22.95 & 22.22 & 71.36 & 30.46 & 46.43 & 23.93 & 8.833 & 4.398 & 4.964 & 5.663 & 20.14 & 14.52 & 24.25 \\
vox2vec$\dagger$  & 38.43 & 43.18 & 20.69 & 20.53 & 33.47 & 72.13 & 36.21 & 66.26 & 42.14 & 18.09 & 8.152 & 10.32 & 14.75 & 37.32 & 29.73 & 32.76 \\
HySparK$\dagger$  & 43.62 & 43.39 & 41.82 & 23.36 & 26.58 & 71.66 & 46.87 & 67.59 & 40.93 & 18.61 & 7.166 & 15.85 & 13.45 & 31.78 & 24.76 & 34.50 \\
VoCo$\dagger$  & \textbf{72.57} & \textbf{78.74} & 31.43 & \underline{79.82} & \underline{43.08} & \underline{46.58} & \underline{85.35} & \textbf{72.39} & \textbf{80.57} & \underline{63.95} & \underline{48.63} & 21.42 & \underline{43.27} & 24.44 & \underline{44.81} & \underline{55.81} \\
\rowcolor{gray!15} \textbf{Hi-End-MAE}  & \underline{71.98} & \underline{74.43} & \textbf{50.71} & \textbf{80.59} & \textbf{45.36} & \textbf{49.97} & 85.08 & \underline{70.52} & \underline{77.40} & \textbf{64.84} & \textbf{54.52} & \textbf{42.48} & 41.22 & \textbf{41.94} & \textbf{54.15} & \textbf{60.35} \\
\hline

\Xhline{1px}
\end{tabular}
}

\vspace{-3mm}
\label{tab:amos}
\end{table*}

\renewcommand{\multirowsetup}{\centering}  
\begin{table*}[t]
\centering
\caption{Comparison of different methods with 1\%, 10\%, and 100\%  proportions on FLARE'22~\citep{flare}. We report the DSC (\%) performance. \textbf{val} (bold) / \underline{val} (underline) : top method / second method. $\dagger$ denotes we utilize official pre-training weights. The abbreviations Liv, RKid, LKid, Spl, Pan, Aor, IVC, RAG, LAG, Gall, Eso, Sto, and Duo correspond to Liver, Right kidney, Left kidney, Spleen, Pancreas, Aorta, Inferior Vena Cava, Right Adrenal Gland, and Left Adrenal Gland, Gallbladder, Esophagus, Stomach, and Duodenum, respectively.}
\vspace{-2mm}
\resizebox{0.99\linewidth}{!}
{
\begin{tabular}{l c c c c c c c c c c c c c c}
\Xhline{1px} 
\multirow{2}{*}{Method} & \multicolumn{13}{c}{FLARE'22 (100\%)}  & \multirow{2}{*}{Avg} \\
\cline{2-14}
& Liv & RKid & LKid & Spl & Pan & Aor & IVC & RAG & LAG & Gall & Eso & Sto & Duo & \\
\hline
\multicolumn{1}{c}{\textit{{\color{Gray} From scratch}}} \\
UNETR & 93.48 & 78.17 & 76.43 & 86.14 & 63.80 & 83.44 & 74.17 & 64.88 & 52.07 & 55.77 & 69.04 & 72.44 & 50.73 & 70.81 \\
SwinUNETR & 94.74 & 81.59 & 79.54 & 87.68 & 69.38 & 90.30 & 80.18 & 67.99 & 61.63 & 57.25 & 70.99 & 78.16 & 60.45 & 75.38 \\

\hline
\multicolumn{1}{c}{\textit{{\color{Gray} General SSL}}} \\
SparK & 96.09 & 84.12 & 85.96 & 91.47 & 79.64 & 92.37 & 83.61 & 70.22 & 68.58 & 62.07 & 75.90 & 85.20 & 73.35 & 80.67 \\
MAE & 96.85 & 88.12 & \underline{87.60} & 92.28 & 82.87 & 93.88 & \underline{86.34} & 74.56 & 69.07 & 64.24 & 76.66 & \underline{88.33} & 72.42 & 82.56 \\

\hline
\multicolumn{1}{c}{\textit{{\color{Gray} Medical SSL}}} \\
MG$\dagger$ & 93.60 & 71.01 & 71.53 & 80.15 & 71.59 & 86.49 & 78.47 &0.00&0.00& 56.39 &0.00& 74.24 & 61.76 & 57.33 \\
TransVW$\dagger$ & 94.20 & 73.84 & 72.08 & 82.39 & 76.07 & 88.91 & 79.30 & 70.18 & 65.60 & 59.71 & 77.46 & 80.57 & 64.69 & 75.78 \\
UniMiSS$\dagger$ & 95.56 & 75.33 & 73.44 & 90.29 & 72.25 & 89.90 & 79.14 & 68.67 & 64.98 & 58.99 & 66.19 & 79.33 & 57.16 & 74.71 \\
SUP$\dagger$ & 94.87 & 81.85 & 80.24 & 87.55 & 69.92 & 89.07 & 78.72 & 66.60 & 58.84 & 57.65 & 72.53 & 77.80 & 58.80 & 74.96 \\
PRLv2$\dagger$ & 83.67 & 68.04 & 62.64 & 74.11 & 71.04 & 84.84 & 74.04 &0.00&0.00& 51.34 &0.00& 78.54 & 57.44 & 54.29 \\
GVSL$\dagger$ & 93.95 & 73.98 & 73.36 & 85.23 & 71.89 & 90.55 & 74.48 & 67.73 & 59.81 & 55.63 & 69.62 & 79.44 & 56.74 & 73.27 \\
vox2vec$\dagger$ & 80.67 & 75.50 & 75.42 & 76.25 & 73.10 & 89.93 & 79.11 & 59.76 & 55.73 & 49.15 & 67.18 & 77.91 & 54.56 & 70.33 \\
HySparK$\dagger$ & 96.45 & \textbf{89.48} & 87.32 & 91.96 & 83.30 & 93.83 & 85.19 & 70.18 & \underline{71.19} & 63.38 & 78.89 & 85.91 & 73.39 & 82.35 \\
VoCo$\dagger$ & \underline{96.89} & 87.76 & 85.82 & \underline{92.62} & \underline{84.32} & \underline{94.29} & 85.29 & \textbf{76.93} & 71.18 & \underline{64.48} & \textbf{80.04} & 87.05 & \underline{73.84} & \underline{83.12} \\
\rowcolor{gray!15} \textbf{Hi-End-MAE}  & \textbf{97.18} & \underline{89.29} & \textbf{88.82} & \textbf{93.41} & \textbf{84.84} & \textbf{94.82} & \textbf{87.88} & \underline{75.85} & \textbf{72.76} & \textbf{65.04} & \underline{79.03} & \textbf{89.48} & \textbf{76.08} & \textbf{84.20} \\

\Xhline{1px} 
 & \multicolumn{13}{c}{FLARE'22 (10\%)}  &  \\
\hline
\multicolumn{1}{c}{\textit{{\color{Gray} From scratch}}} \\
UNETR & 88.78 & 63.45 & 62.53 & 61.20 & 48.54 & 78.70 & 58.17 & 53.17 & 33.78 & 38.22 & 60.25 & 53.91 & 33.35 & 56.46 \\
SwinUNETR & 90.69 & 65.45 & 58.83 & 74.18 & 57.45 & 82.37 & 66.39 & 59.35 & 50.36 & 48.64 & 65.67 & 58.38 & 46.16 & 63.38 \\

\hline
\multicolumn{1}{c}{\textit{{\color{Gray} General SSL}}} \\
SparK & 94.07 & 74.01 & 74.38 & 84.75 & 73.61 & 89.80 & 76.68 & 60.29 & 57.00 & 51.34 & 68.88 & 72.17 & 55.54 & 71.74 \\
MAE & 95.20 & 76.51 & 78.72 & 86.12 & 79.22 & 92.96 & 82.20 & 67.23 & 62.15 & \underline{60.19} & 73.32 & 79.07 & \underline{68.17} & 77.01 \\

\hline
\multicolumn{1}{c}{\textit{{\color{Gray} Medical SSL}}} \\
MG$\dagger$ & 88.57 & 57.05 & 54.98 & 70.38 & 55.78 & 84.82 & 66.29 &0.00&0.00& 44.33 &0.00& 59.50 & 44.65 & 48.18 \\
TransVW$\dagger$ & 89.12 & 53.34 & 51.98 & 62.48 & 63.55 & 83.74 & 70.33 & 58.24 & 46.88 & 47.35 & 66.34 & 64.35 & 49.08 & 62.07 \\
UniMiSS$\dagger$ & 90.40 & 68.55 & 61.09 & 68.87 & 55.41 & 82.96 & 63.74 & 56.52 & 50.02 & 44.89 & 57.56 & 56.12 & 36.73 & 60.99 \\
SUP$\dagger$ & 89.69 & 63.82 & 55.83 & 63.15 & 55.51 & 81.64 & 66.68 & 55.83 & 40.97 & 48.45 & 64.63 & 56.79 & 41.44 & 60.35 \\
PRLv2$\dagger$ & 89.39 & 56.33 & 47.00 & 58.86 & 51.86 & 84.18 & 68.90 &0.00&0.00&0.00&0.00& 64.37 & 37.65 & 42.97 \\
GVSL$\dagger$  & 90.40 & 59.04 & 58.52 & 71.19 & 53.54 & 80.32 & 62.90 & 53.12 & 38.90 & 48.11 & 56.05 & 60.39 & 41.45 & 59.54 \\
vox2vec$\dagger$  & 90.60 & 65.40 & 56.22 & 65.36 & 61.83 & 81.13 & 69.34 & 54.50 & 41.50 & 49.75 & 64.28 & 60.11 & 45.84 & 61.99 \\
HySparK$\dagger$  & 94.59 & 75.05 & 74.70 & 87.15 & 73.25 & 90.84 & 77.03 & 69.03 & 59.52 & 53.80 & 74.05 & 69.31 & 58.41 & 73.60 \\
VoCo$\dagger$  & \underline{96.37} & \underline{80.70} & \underline{81.24} & \underline{89.06} & \underline{80.28} & \underline{93.28} & \underline{82.45} & \textbf{72.89} & \textbf{67.15} & 59.73 & \textbf{77.75} & \underline{80.02} & 63.89 & \underline{78.84} \\
\rowcolor{gray!15} \textbf{Hi-End-MAE}  & \textbf{96.53} & \textbf{84.60} & \textbf{82.82} & \textbf{92.67} & \textbf{82.40} & \textbf{93.85} & \textbf{85.25} & \underline{69.57} & \underline{65.17} & \textbf{61.90} & \underline{76.39} & \textbf{83.74} & \textbf{72.64} & \textbf{80.58} \\

\Xhline{1px} 
 & \multicolumn{13}{c}{FLARE'22 (1\%)}  &  \\
\hline
\multicolumn{1}{c}{\textit{{\color{Gray} From scratch}}} \\
UNETR  & 79.55 & 33.05 & 21.85 & 45.32 & 14.27 & 46.34 & 27.78 & 4.215 &0.00& 20.79 & 24.23 & 20.49 & 3.979 & 26.30 \\
SwinUNETR  & 80.45 & 45.22 & 37.37 & 45.01 & 21.44 & 63.45 & 45.46 & 13.36 & 7.671 & 31.44 & 27.47 & 31.90 & 16.26 & 35.89 \\

\hline
\multicolumn{1}{c}{\textit{{\color{Gray} General SSL}}} \\
SparK  & 83.63 & 41.92 & 32.04 & 48.15 & 22.89 & 58.12 & 44.64 & 9.142 & 13.18 & 23.97 & 33.74 & 43.18 & 19.62 & 36.48 \\
MAE  & 88.87 & \textbf{72.24} & \underline{59.91} & 70.86 & \textbf{61.85} & \textbf{88.75} & \underline{70.75} & \textbf{44.77} & \textbf{46.96} & 34.47 & \underline{59.08} & \underline{66.80} & \underline{45.21} & \underline{62.35} \\

\hline
\multicolumn{1}{c}{\textit{{\color{Gray} Medical SSL}}} \\
MG$\dagger$  & 72.64 & 33.54 & 34.72 & 39.79 & 16.10 & 61.64 & 38.19 &0.00&0.00& 20.60 &0.00& 29.48 & 8.115 & 27.30 \\
TransVW$\dagger$  & 62.50 &0.00&0.00&0.00&0.00&0.00&0.00&0.00&0.00&0.00&0.00&0.00&0.00& 4.807 \\
UniMiSS$\dagger$  & 72.90 & 35.59 & 24.13 & 22.31 & 19.10 & 40.34 & 21.63 &0.00& 5.034 & 24.43 & 15.66 & 33.80 & 9.015 & 24.92 \\
SUP$\dagger$  & 83.17 & 38.20 & 34.70 & 52.42 & 20.88 & 59.71 & 39.53 & 5.548 & 10.33 & 25.23 & 23.94 & 31.95 & 12.67 & 33.72 \\
PRLv2$\dagger$  & 80.31 & 33.82 & 20.32 & 46.03 & 18.92 & 65.71 & 43.38 &0.00&0.00& 20.29 &0.00& 25.18 & 6.190 & 27.71 \\
GVSL$\dagger$  & 75.57 & 26.34 & 26.64 & 41.81 & 14.15 & 47.10 & 24.30 & 2.690 &0.00& 22.09 & 18.12 & 28.99 & 14.45 & 26.33 \\
vox2vec$\dagger$ & 69.73 & 33.29 & 33.66 & 37.72 & 29.60 & 58.24 & 46.40 & 6.671 & 8.964 & 19.83 & 40.53 & 35.47 & 23.25 & 34.11 \\
HySparK$\dagger$  & 83.07 & 43.34 & 37.74 & 51.38 & 22.88 & 66.24 & 47.03 & 15.03 & 4.253 & 31.27 & 25.19 & 37.11 & 23.47 & 37.54 \\
VoCo$\dagger$  & \textbf{91.21} & 59.29 & \textbf{66.63} & \textbf{78.29} & 50.82 & 79.31 & 66.23 & 31.61 & 35.00 & \textbf{41.35} & 50.40 & 62.52 & 36.85 & 57.66 \\
\rowcolor{gray!15} \textbf{Hi-End-MAE} & \underline{90.29} & \underline{71.13} & 59.75 & \underline{71.77} & \underline{61.66} & \underline{86.03} & \textbf{74.62} & \underline{43.34} & \textbf{46.96} & \underline{34.97} & \textbf{59.54} & \textbf{71.26} & \textbf{50.61} & \textbf{63.22} \\

\hline

\Xhline{1px}
\end{tabular}
}

\vspace{-3mm}
\label{tab:flare}
\end{table*}

\section{Visualization}
\noindent\textbf{Reconstruction visualization.} We visualize 3D reconstruction results to check what Hi-End-MAE
learns in pre-training with different mask ratios. As shown in Fig.\ref{fig:recon}, our method can almost reconstruct the different shapes of organs, bones, and other details from the very small portion of unmasked patches.

\noindent\textbf{More segmentation result visualization.} 
Visualization results on BTCV~\citep{btcv}, CT-ORG~\citep{ctorg}, Sliver07~\citep{sliver07}, WORD~\citep{word}, AMOS~\citep{amos}, and FLARE’22~\citep{flare} are shown in Fig.\ref{fig:vis}. It can be observed that, compared to training from scratch, the high-quality representations learned by Hi-End-MAE provide significant benefits to downstream tasks.

\end{document}